\documentclass[10pt,twocolumn,letterpaper]{article}

\usepackage{cvpr}      

\usepackage[dvipsnames]{xcolor}

\definecolor{cvprblue}{rgb}{0.21,0.49,0.74}
\usepackage[pagebackref,breaklinks,colorlinks,citecolor=cvprblue]{hyperref}
\usepackage{booktabs}
\usepackage{multirow}
\usepackage{amssymb}
\usepackage[normalem]{ulem}
\usepackage{hyperref}
\def\paperID{9677} 
\def\confName{CVPR}
\def\confYear{2024}

\title{Revisiting Adversarial Training under Long-Tailed Distributions}

\author{
Xinli Yue, Ningping Mou, Qian Wang, Lingchen Zhao\thanks{Corresponding author.} \\
Key Laboratory of Aerospace Information Security and Trusted Computing, Ministry of Education,\\
School of Cyber Science and Engineering, Wuhan University, Wuhan 430072, China\\
\tt\small\{xinliyue,ningpingmou,qianwang,lczhaocs\}@whu.edu.cn \\
}

\begin{document}
\maketitle
\begin{abstract}
Deep neural networks are vulnerable to adversarial attacks, often leading to erroneous outputs. Adversarial training has been recognized as one of the most effective methods to counter such attacks. However, existing adversarial training techniques have predominantly been tested on balanced datasets, whereas real-world data often exhibit a long-tailed distribution, casting doubt on the efficacy of these methods in practical scenarios.

In this paper, we delve into adversarial training under long-tailed distributions. Through an analysis of the previous work ``RoBal'', we discover that utilizing Balanced Softmax Loss alone can achieve performance comparable to the complete RoBal approach while significantly reducing training overheads. Additionally, we reveal that, similar to uniform distributions, adversarial training under long-tailed distributions also suffers from robust overfitting. To address this, we explore data augmentation as a solution and unexpectedly discover that, unlike results obtained with balanced data, data augmentation not only effectively alleviates robust overfitting but also significantly improves robustness. We further investigate the reasons behind the improvement of robustness through data augmentation and identify that it is attributable to the increased diversity of examples. Extensive experiments further corroborate that data augmentation alone can significantly improve robustness. Finally, building on these findings, we demonstrate that compared to RoBal, the combination of BSL and data augmentation leads to a +6.66\% improvement in model robustness under AutoAttack on CIFAR-10-LT. Our code is available at: \href{https://github.com/NISPLab/AT-BSL}{https://github.com/NISPLab/AT-BSL}.

\end{abstract}
\section{Introduction}
\label{sec:intro}

\begin{figure}[t]
	\centering
	\includegraphics[width=1.0\linewidth]{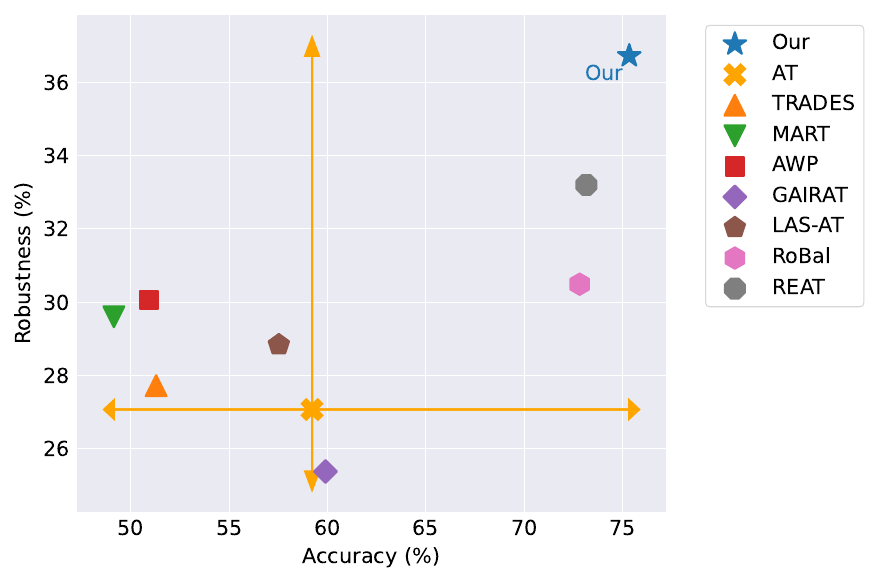}
        \vspace{-0.2 in}
	\caption{The clean accuracy and robustness under AutoAttack (AA)~\cite{croce2020reliable} of various adversarial training methods using WideResNet-34-10~\cite{zagoruyko2016wide} on CIFAR-10-LT~\cite{krizhevsky2009learning}. Our method, building upon AT~\cite{madry2018towards} and BSL~\cite{ren2020balanced}, leverages data augmentation to improve robustness, achieving a +6.66\% improvement over the SOTA method RoBal~\cite{wu2021adversarial}. REAT~\cite{li2023adversarial} is a concurrent work with ours, yet to be published.}
	\label{fig_acc_rob_wrn_cifar10}
	\vspace{-0.2 in}
\end{figure}

It is well-known that deep neural networks (DNNs) are vulnerable to adversarial attacks, where attackers can induce errors in DNNs' recognition results by adding perturbations that are imperceptible to the human eye~\cite{szegedy2013intriguing,goodfellow2014explaining}. Many researchers have focused on defending against such attacks. Among the various defense methods proposed, adversarial training is recognized as one of the most effective approaches. It involves integrating adversarial examples into the training set to enhance the model's generalization capability against these examples~\cite{madry2018towards,zhang2019theoretically,wang2019improving,wu2020adversarial,zhang2020geometry,jia2022adversarial}. In recent years, significant progress has been made in the field of adversarial training. However, we note that almost all studies on adversarial training utilize balanced datasets like CIFAR-10, CIFAR-100~\cite{krizhevsky2009learning}, and Tiny-ImageNet~\cite{le2015tiny} for performance evaluation. In contrast, real-world datasets often exhibit an imbalanced, typically long-tailed distribution. Hence, the efficacy of adversarial training in practical systems should be reassessed using long-tailed datasets~\cite{van2018inaturalist,gupta2019lvis}. 

To the best of our knowledge, RoBal~\cite{wu2021adversarial} is the sole published work that investigates the adversarial robustness under the long-tailed distribution. However, due to its complex design, RoBal demands extensive training time and GPU memory, which somewhat limits its practicality. Upon revisiting the design and principles of RoBal, we find that its most critical component is the Balanced Softmax Loss (BSL)~\cite{ren2020balanced}. We observe that combining AT~\cite{madry2018towards} with BSL to form AT-BSL can match RoBal's effectiveness while significantly reducing training overhead. Following Occam's Razor principle, where entities should not be multiplied without necessity~\cite{jefferys1992ockham}, we advocate using AT-BSL as a substitute for RoBal. In this paper, we base our studies on AT-BSL.

In the course of our study on the robustness of models under long-tailed distribution, we encounter another significant finding: adversarial training with long-tailed distribution data, similar to training on balanced datasets, also leads to the issue of robust overfitting~\cite{rice2020overfitting}. Previous works on balanced datasets often employed data augmentation to mitigate this robust overfitting~\cite{rice2020overfitting,wu2020adversarial,carmon2019unlabeled,gowal2020uncovering,rebuffi2021data}. Hence, a straightforward approach is to attempt the use of data augmentation to alleviate the robust overfitting issue in adversarial training with long-tailed distribution. Our results align with findings on balanced datasets, indicating that data augmentation can mitigate robust overfitting. However, contrary to findings on balanced datasets where it was concluded that data augmentation alone can not improve robustness~\cite{rice2020overfitting,wu2020adversarial,rebuffi2021data}, we find that data augmentation techniques, including MixUp~\cite{zhang2018mixup}, Cutout~\cite{devries2017improved}, CutMix~\cite{yun2019cutmix}, AugMix~\cite{hendrycks2019augmix}, AutoAugment (AuA)~\cite{cubuk2019autoaugment}, RandAugment (RA)~\cite{cubuk2020randaugment} and TrivialAugment (TA) ~\cite{muller2021trivialaugment}, can significantly improve robustness. Hence, we introduce the following query: Why does data augmentation improve robustness? We hypothesize that data augmentation augments example diversity, enabling the model to learn richer representations thereby improving its robustness. Subsequently, we validate our hypothesis through ablation studies on RA.

Our contributions are summarized as follows:
\begin{itemize}
    \item Through ablation studies, we discover that BSL is the most critical component of RoBal, and the streamlined method AT-BSL significantly reduces training time and memory usage compared to RoBal.

    \item We observe that data augmentation not only mitigates robust overfitting in adversarial training under long-tailed distributions but also substantially improves robustness.

    \item We propose a hypothesis about the reasons for data augmentation improving robustness and validate this hypothesis through experiments.

    \item Comprehensive empirical evidence demonstrates that our discoveries generalize across various data augmentation strategies, model architectures, and datasets.
    
\end{itemize}

\section{Related Works}
\noindent \textbf{Long-Tailed Recognition.} Long-tailed distributions refer to a common imbalance in training set where a small portion of classes (head) have massive examples, while other classes (tail) have very few examples~\cite{van2018inaturalist,gupta2019lvis}. Models trained under such distribution tend to exhibit a bias towards the head classes, resulting in poor performance for the tail classes. Traditional rebalancing techniques aim at addressing the long-tailed recognition problem include re-sampling~\cite{shen2016relay,Kang2020Decoupling,wang2020devil,zhang2021learning} and cost-sensitive learning~\cite{cui2019class,lin2017focal}, which often improve the performance of tail classes at the expense of head classes. To mitigate these adverse effects, some methods handle class-specific attributes through perspectives such as margins~\cite{wang2017learning} and biases~\cite{ren2020balanced}. Recently, more advanced techniques like class-conditional sharpness-aware minimization~\cite{zhou2023class}, feature clusters compression~\cite{li2023fcc}, and global-local mixture consistency cumulative learning~\cite{du2023global} have been introduced, further improving the performance of long-tailed recognition. However, these works have been devoted to improving clean accuracy, and investigations into the adversarial robustness of long-tailed recognition remain scant.

\noindent \textbf{Adversarial Training.} The philosophy of adversarial training involves integrating adversarial examples into the training set, thereby improving the model’s generalizability to such examples. Adversarial training addresses a min-max problem, with the inner maximization dedicated to generating the strongest adversarial examples and the outer minimization aimed at optimizing the model parameters. The quintessential method of adversarial training is AT~\cite{madry2018towards}, which can be mathematically represented as follows:
\begin{equation}
\begin{aligned}
&\underset{\theta_m}{\operatorname{argmin}} \mathcal{L}_{\min }\left(\theta_m ; x^{\prime}, y\right), \\ 
&\text{ where } x^{\prime}=\underset{\left\|x^{\prime}-x\right\|_p \leq \epsilon}{\operatorname{argmax}} \mathcal{L}_{\max }\left(\theta_m ; x^{\prime}, y\right). 
\end{aligned}
\end{equation}
where $x^{\prime}$ is an adversarial example constrained by $\ell_p$ norm for clean examples $x$, $y$ is the label of $x$, $\theta_m$ is the parameter of the model $m$, $\epsilon$ is the perturbation size, $\mathcal{L}_{\max }$ is the internal maximization loss, and $\mathcal{L}_{\min }$ is the external minimization loss.

Building upon the foundation of AT~\cite{madry2018towards}, subsequent works developed advanced adversarial training techniques such as TRADES~\cite{zhang2019theoretically}, MART~\cite{wang2019improving}, AWP~\cite{wu2020adversarial}, GAIRAT~\cite{zhang2020geometry}, and LAS-AT~\cite{jia2022adversarial}. However, these adversarial training methods were predominantly experimented with on balanced datasets like CIFAR-10 and CIFAR-100.

\noindent \textbf{Robustness under Long-Tailed Distribution.} Previous adversarial training works were concentrated mainly on balanced datasets. However, data in the real world are seldom balanced; they are more commonly characterized by long-tailed distributions~\cite{van2018inaturalist,gupta2019lvis}. Therefore, a critical criterion for assessing the practical utility of adversarial training should be its performance on long-tailed distributions. To our knowledge, RoBal~\cite{wu2021adversarial} is the only work that investigates adversarial training on long-tailed datasets. In Section \ref{analyze_robal}, we delve into the components of RoBal, improving the efficacy of long-tailed adversarial training based on our findings. Moreover, some works~\cite{xu2021robust,ma2022tradeoff,wei2023cfa,yue2023revisiting} have already indicated that adversarial training on balanced datasets can lead to significant robustness disparities across classes. Whether this disparity is exacerbated on long-tailed datasets remains an open question for further exploration.

\noindent \textbf{Data Augmentation.} In standard training regimes, data augmentation has been validated as an effective tool to mitigate overfitting and improve model generalization, regardless of whether the data distribution is balanced or long-tailed~\cite{zhang2021bag,xu2021towards,ahn2023cuda,du2023global}. The most commonly utilized augmentation techniques for image classification tasks include random flips, rotations, and crops~\cite{he2016deep}. More sophisticated augmentation methods like MixUp~\cite{zhang2018mixup}, Cutout~\cite{devries2017improved}, and CutMix~\cite{yun2019cutmix} have been shown to yield superior results in standard training contexts. Furthermore, augmentation strategies such as Augmix~\cite{hendrycks2019augmix}, AuA~\cite{cubuk2019autoaugment}, RA~\cite{cubuk2020randaugment}, and TA~\cite{muller2021trivialaugment}, which employ a learned or random combination of multiple augmentations, have elevated the efficacy of data augmentation to new heights, heralding the advent of the era of automated augmentation.

\section{Analysis of RoBal}
\label{analyze_robal}
\subsection{Preliminaries}
RoBal~\cite{wu2021adversarial}, in comparison to AT~\cite{madry2018towards}, incorporates four additional components: 1) cosine classifier; 2) Balanced Softmax Loss~\cite{ren2020balanced}; 3) class-aware margin; 4) TRADES regularization~\cite{zhang2019theoretically}.

\noindent \textbf{Cosine Classifier.} In basic classification tasks employing a standard linear classifier, the predicted logit for class $i$ can be represented as:
\begin{equation}
\begin{aligned}
g(f(x))_i&=W_i^T f(x)+b_i\\
&=\left\|W_i\right\| \cdot\|f(x)\| \cos \theta_i+b_i\\
&=z_i+b_i, 
\end{aligned}
\end{equation}
where $g(\cdot)$ is the liner classifer. In this formulation, the prediction depends on three factors: 1) the magnitude of the weight vector $\left\|W_i\right\|$ and the feature vector $\|f(x)\|$; 2) the angle between them, expressed as $\cos \theta_i$; and 3) the bias of the classifier $b_i$.

The above decomposition illustrates that simply by scaling the norm of examples in feature space, the predictions of the examples can be altered. In linear classifiers, the scale of the weight vector $\left\|W_i\right\|$ often diminishes in tail classes, thereby impacting the recognition performance for tail classes. Consequently, ~\cite{wu2021adversarial} endeavors to utilize a cosine classifier~\cite{pang2020boosting} to mitigate the scale effects of features and weights. And in the cosine classifier, the predicted logit for class $i$ can be represented as:
\begin{equation}
\begin{aligned}
h(f(x))_i&=s \cdot\left(\frac{W_i^T f(x)}{\left\|W_i\right\|\|f(x)\|}\right)+b_i\\
&=s \cdot \cos \theta_i+b_i,
\end{aligned}
\end{equation}
where $h(\cdot)$ is the cosine classifier, $\|\cdot\|$ denotes the $\ell_2$ norm of the vector, $s$ is the scaling factor.

\noindent \textbf{Balanced Softmax Loss.} An intuitive and widely adopted approach to addressing class imbalance is to assign class-specific biases during training for the cross-entropy (CE) loss. ~\cite{wu2021adversarial} employs the formulation by ~\cite{ren2020balanced,menon2020long}, denoted as $b_i=\tau_b \log \left(n_i\right)$, where the modified cross-entropy loss, namely Balanced Softmax Loss (\textbf{BSL}), becomes:
\begin{equation}
\begin{aligned}
&\mathcal{L}_0(h(f(x)), \mathrm{y})=-\log \left(\frac{e^{s \cdot \cos \theta_y+b_y}}{\sum_i e^{s \cdot \cos \theta_i+b_i}}\right)\\
&=\log \left(1+\sum_{i \neq y} e^{s \cdot\left(\cos \theta_i-\cos \theta_y\right)+\tau_b \log \left(\frac{n_i}{n_y}\right)}\right),
\end{aligned}
\end{equation}
where $n_i$ is the number of examples in the $i$-th class, and $\tau_b$ is a hyperparameter controling the calculation of bias. BSL adapts to the label distribution shift between training and testing by adding specific biases to each class based on the number of examples in each class to improve long-tailed recognition performance~\cite{ren2020balanced}.

\noindent \textbf{Class-Aware Margin.} However, when considering the margin representation, the margin from the true class $y$ to class $i$, denoted by $\tau_b \log \left({n_i}/{n_y}\right)$, becomes negative when $n_y>n_i$, leading to poorer discriminative representations and classifier learning for head classes. To address this, ~\cite{wu2021adversarial} introduces a class-aware margin term~\cite{pang2020boosting}, which assigns a larger margin value to head classes as compensation:
\begin{equation}
m_i=\frac{\tau_m}{s} \log \frac{n_i}{n_{\min }}+m_0.
\end{equation}
The first term increases with $n_i$ and reaches its minimum of zero when $n_i=n_{\min }$, with $\tau_m$ as the hyperparameter controlling the trend. The second term, $m_0>0$, is a uniform boundary for all classes, a common strategy in networks based on cosine classifiers. Add this class-aware margin $m_i$ to $\mathcal{L}_0$ to become $\mathcal{L}_1$:
\begin{equation}
\begin{aligned}
&\mathcal{L}_1(h(f(x)), \mathrm{y})\\
&=-\log \left(\frac{e^{s\left(\cos \theta_y-m_y\right)+b_y}}{e^{s\left(\cos \theta_y-m_y\right)+b_y}+\sum_{i \neq y} e^{s \cos \theta_i+b_i}}\right).
\end{aligned}
\end{equation}

\begin{table*}[t]
\begin{center}
\caption{The clean accuracy, robustness, time (average per epoch) and memory (GPU) using ResNet-18~\cite{he2016deep} on CIFAR-10-LT following the integration of components from RoBal~\cite{wu2021adversarial} into AT~\cite{madry2018towards}. The best results are \textbf{bolded}. The second best results are \uline{underlined}. Cos: Cosine Classifier; BSL: Balanced Softmax Loss~\cite{ren2020balanced}; CM: Class-aware Margin~\cite{wu2021adversarial}; TRADES: TRADES Regularization~\cite{zhang2019theoretically}.}
\label{tab_robal_ablation}
\vspace{-0.1 in}
\setlength{\tabcolsep}{1.0mm}{\begin{tabular}{ccccccccccccc}
\toprule[1.0pt]
\multirow{2}{*}{Method} & \multicolumn{4}{c}{Components} & \multicolumn{6}{c}{Accuracy}                                                                        & \multicolumn{2}{c}{Efficiency} \\ \cmidrule(l){2-5} \cmidrule(l){6-11} \cmidrule(l){12-13}
                        & Cos   & BSL   & CM   & TRADES  & Clean          & FGSM           & PGD            & CW             & LSA            & AA             & Time (s)            & Memory (MiB)       \\ \midrule
AT~\cite{madry2018towards}                     &       &       &      &         & 54.91          & 32.21          & 28.05          & 28.28          & 28.73          & 26.75          & \uline{21.36}           & \textbf{946} \\
AT-BSL                 &       & \checkmark     &      &         & 70.21          & 37.44          & 31.91          & \textbf{31.45} & \textbf{32.25} & \uline{29.48}    & \textbf{21.00}  & \textbf{946} \\
AT-BSL-Cos             & \checkmark     & \checkmark     &      &         & \textbf{71.99} & 39.41          & 34.73          & 30.27          & 29.94          & 28.43          & 22.39           & \textbf{946} \\
AT-BSL-Cos-TRADES      & \checkmark     & \checkmark     &      & \checkmark       & 69.31          & \uline{39.62}    & \uline{34.87}    & 30.19          & 30.15          & 28.64          & 38.91           & \uline{1722}         \\
RoBal~\cite{wu2021adversarial}                 & \checkmark     & \checkmark     & \checkmark    & \checkmark       & \uline{70.34}    & \textbf{40.50} & \textbf{35.93} & \uline{31.05}    & \uline{31.10}    & \textbf{29.54} & 39.03           & \uline{1722}         \\ \bottomrule[1.0pt]
\end{tabular}}
\end{center}
\vspace{-0.1 in}
\end{table*}

\noindent \textbf{TRADES Regularization.} ~\cite{wu2021adversarial} incorporates a KL regularization term following TRADES~\cite{zhang2019theoretically}, thereby modifying the overall loss function to:
\begin{equation}
\mathcal{L}_{\text {min }}=\mathcal{L}_1\left(h\left(f\left(x^{\prime}\right)\right), \mathrm{y}\right)+\beta \cdot \mathrm{KL}\left(h\left(f\left(x^{\prime}\right)\right), h(f(x))\right),
\end{equation}
where $\beta$ serves as a hyperparameter to control the intensity of the TRADES regularization.

\subsection{Ablation Studies of RoBal}
To investigate the role of each component in RoBal~\cite{wu2021adversarial}, we conduct ablation studies on it. Specifically, we incrementally add each component of RoBal to AT~\cite{madry2018towards} and then evaluate the method's clean accuracy, robustness, training time per epoch, and memory usage. The results are summarized in Table \ref{tab_robal_ablation}. Note that the parameters utilized in Table \ref{tab_robal_ablation} adhere strictly to the default settings of ~\cite{wu2021adversarial}, and the details about adversarial attacks are in Section \ref{subsection_settings}. We observe that the AT-BSL method outperforms AT~\cite{madry2018towards} in terms of clean accuracy and adversarial robustness. However, upon integrating a cosine classifier with AT-BSL, while the robustness under PGD~\cite{madry2018towards} significantly improves, robustness under adaptive attacks like CW~\cite{carlini2017towards}, LSA~\cite{hitaj2021evaluating}, and AA~\cite{croce2020reliable} notably decreases. This aligns with observations in REAT~\cite{li2023adversarial}, suggesting that the cosine classifier (scale-invariant classifier) used in RoBal may lead to gradient vanishing when generating adversarial examples with cross-entropy loss. This is attributed to the normalization of weights and features in the classification layer, which substantially reduces the gradient scale, impeding the generation of potent adversarial examples~\cite{li2023adversarial}. Further additions of TRADES regularization~\cite{zhang2019theoretically} and class-aware margin do not yield substantial improvements in robustness under AA, yet markedly increase training time and memory consumption. In fact, AT-BSL alone can match the complete RoBal in terms of clean accuracy and robustness under AA. Therefore, in line with Occam's Razor~\cite{jefferys1992ockham}, we advocate using AT-BSL, which renders adversarial training more efficient without sacrificing significant performance. The $\mathcal{L}_{min}$ formula of AT-BSL is as follows:
\begin{equation}
\label{eq_at_bsl}
\begin{aligned}
\mathcal{L}_{\text {min }}&=\mathcal{L}_0\left(g\left(f\left(x^{\prime}\right)\right), \mathrm{y}\right)\\
&=-\log \left(\frac{e^{z_y+b_y}}{\sum_i e^{z_i+b_i}}\right)\\
&=-\log \left(\frac{n_y^{\tau_b} \cdot e^{z_y}}{\sum_i n_i^{\tau_b} \cdot e^{z_i}}\right).
\end{aligned}
\vspace{-0.1 in}
\end{equation}

\begin{figure}[t]
    \centering
    \begin{tabular}{cc}
        \includegraphics[width=0.45\linewidth]{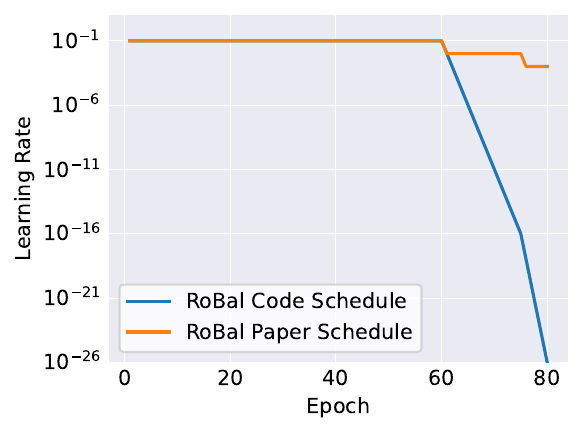} &
        \includegraphics[width=0.45\linewidth]{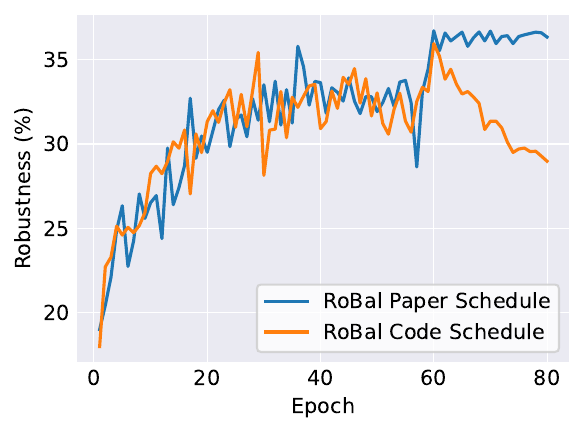}
        \\
        (a) & (b) 
    \end{tabular}
    \vspace{-0.1 in}
    \caption{Learning rate scheduling analysis of RoBal~\cite{wu2021adversarial}. (a) comparison of the learning rate schedules: `RoBal Code Schedule’ from the source code and `RoBal Paper Schedule’ as described in the publication. (b) the evolution of test robustness under PGD-20~\cite{madry2018towards} using ResNet-18 on CIFAR-10-LT across training epochs.}
    \label{fig_robal_over_fitting}
    \vspace{-0.15 in}
\end{figure}

\subsection{Robust Overfitting and Unexpected Discovery}
\noindent \textbf{Discrepancy in Learning Rate Scheduling: Paper Description vs. Code Implementation.} RoBal~\cite{wu2021adversarial} asserts that early stopping is not employed, and the results reported are from the final epoch, the 80th epoch. The declared learning rate schedule is an initial rate of 0.1, with decays at the 60th and 70th epochs, each by a factor of 0.1. After executing the source code of RoBal, we observe, as depicted by the blue line in Fig. \ref{fig_robal_over_fitting}(b), that test robustness remains essentially unchanged after the first learning rate decay (60th epoch), indicating an absence of robust overfitting. It is well-known that adversarial training on CIFAR-10 exhibits significant robust overfitting~\cite{rice2020overfitting}, and given that CIFAR-10-LT has less data than CIFAR-10, the absence of robust overfitting on CIFAR-10-LT is contradictory to the assertion that additional data can alleviate robust overfitting in~\cite{rebuffi2021data}. 

Upon a meticulous examination of the official code provided by RoBal~\cite{wu2021adversarial}, we discover inconsistencies between the implemented learning rate schedule and what is claimed in the paper. 
The official code uses a learning rate schedule starting at 0.1, with a decay of 0.1 per epoch after the 60th epoch and 0.01 per epoch after the 75th epoch (the blue line in Fig. \ref{fig_robal_over_fitting}(a)). This leads to a learning rate as low as 1e-26 by the 80th epoch, potentially limiting learning after the 60th epoch and contributing to the similar performance of models at the 60th and 80th epochs as shown in Fig. \ref{fig_robal_over_fitting}(b).

Subsequently, we adjust the learning rate schedule to what is declared in~\cite{wu2021adversarial} (the orange line in Fig. \ref{fig_robal_over_fitting}(a)) and redraw the robustness curve, represented by the orange line in Fig. \ref{fig_robal_over_fitting}(b). Post-adjustment, a continuous decline in test robustness following the first learning rate decay is observed, aligning with the robust overfitting phenomenon typically seen on CIFAR-10.

Therefore, adversarial training under long-tailed distributions exhibits robust overfitting, similar to balanced distributions. So, how might we resolve this problem? Several works~\cite{rice2020overfitting,wu2020adversarial,carmon2019unlabeled,gowal2020uncovering,rebuffi2021data} have attempted to use data augmentation to alleviate robust overfitting on balanced datasets. 

\begin{figure}[t]
    \centering
    \begin{tabular}{cc}
        \includegraphics[width=0.45\linewidth]{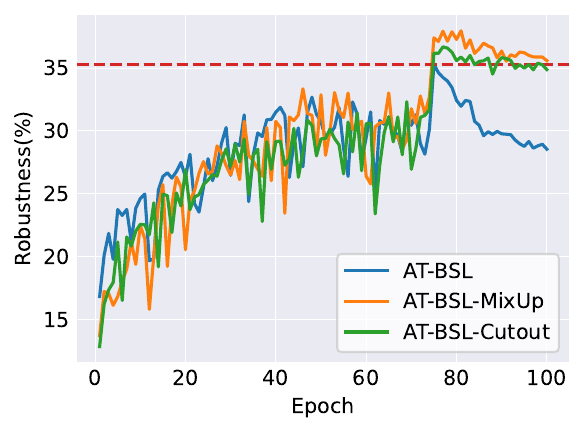} &
        \includegraphics[width=0.45\linewidth]{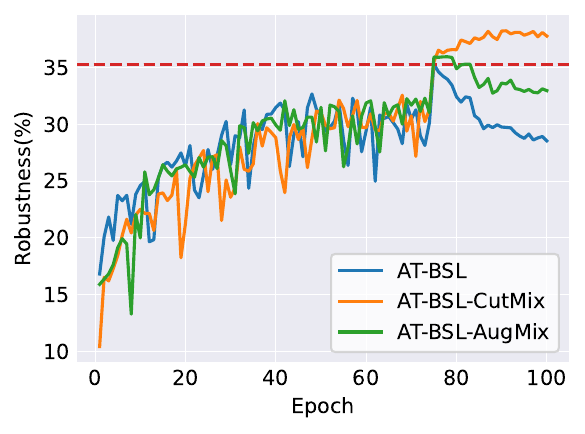} \\
        (a) & (b) \\
        \includegraphics[width=0.45\linewidth]{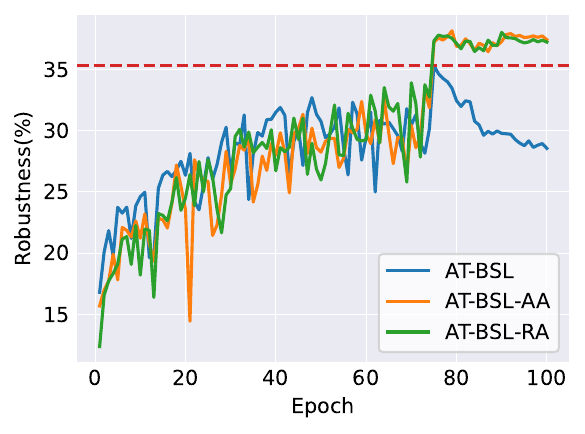} &
        \includegraphics[width=0.45\linewidth]{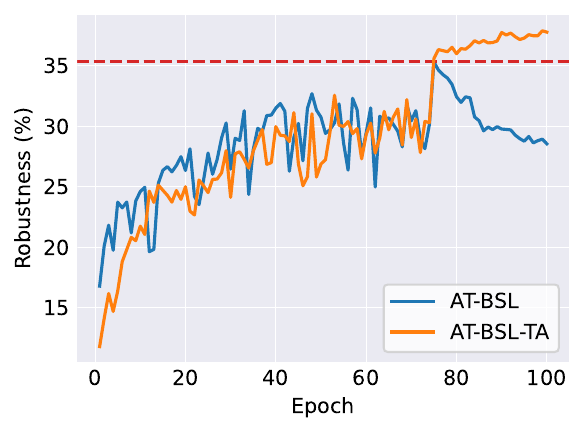} \\
        (c) & (d) \\
    \end{tabular}
    \vspace{-0.1 in}
    \caption{The evolution of test robustness under PGD-20 using ResNet-18 on CIFAR-10-LT for AT-BSL using different data augmentation strategies across training epochs. For reference, the red dashed lines in each panel represent the robustness of the best checkpoint of AT-BSL. Due to the density of the illustrations, the results have been compartmentalized into four distinct panels: (a), (b), (c), and (d).}
    \label{fig_aug}
\vspace{-0.2 in}
\end{figure}

\noindent \textbf{Testing MixUp.} ~\cite{rice2020overfitting,wu2020adversarial,rebuffi2021data} suggest that on CIFAR-10, MixUp~\cite{zhang2018mixup} can alleviate robust overfitting. Therefore, we posit that on the long-tailed version of CIFAR-10, CIFAR-10-LT, MixUp would also mitigate robust overfitting. In Fig. \ref{fig_aug}(a), it is evident that AT-BSL-MixUp, which utilizes MixUp, significantly alleviates robust overfitting compared to AT-BSL. Furthermore, we unexpectedly discover that MixUp markedly improves robustness. This observation is inconsistent with previous findings in balanced datasets~\cite{rice2020overfitting,wu2020adversarial,rebuffi2021data}, where it was concluded that data augmentation alone does not improve robustness.

\noindent \textbf{Exploring data augmentation.} Following the validation of the MixUp hypothesis, our investigation expands to assess whether other augmentation techniques could alleviate robust overfitting and improve robustness. This includes augmentations like Cutout~\cite{devries2017improved}, CutMix~\cite{yun2019cutmix}, AugMix~\cite{hendrycks2019augmix}, TA~\cite{muller2021trivialaugment}, AuA~\cite{cubuk2019autoaugment}, and RA~\cite{cubuk2020randaugment}. Analogous to our analysis of MixUp, we report the robustness achieved by these augmentation techniques during training in Fig. \ref{fig_aug}. Firstly, our findings indicate that each augmentation technique mitigated robust overfitting, with CutMix, AuA, RA, and TA exhibiting almost negligible instances of this phenomenon. Furthermore, we observe that robustness attained by each augmentation surpasses that of the vanilla AT-BSL, further corroborating that \uline{data augmentation alone can improve robustness}.

\section{Why Data Augmentation Can Improve Robustness}

\begin{figure}[t]
    \centering
    \begin{tabular}{cc}
        \includegraphics[width=0.465\linewidth]{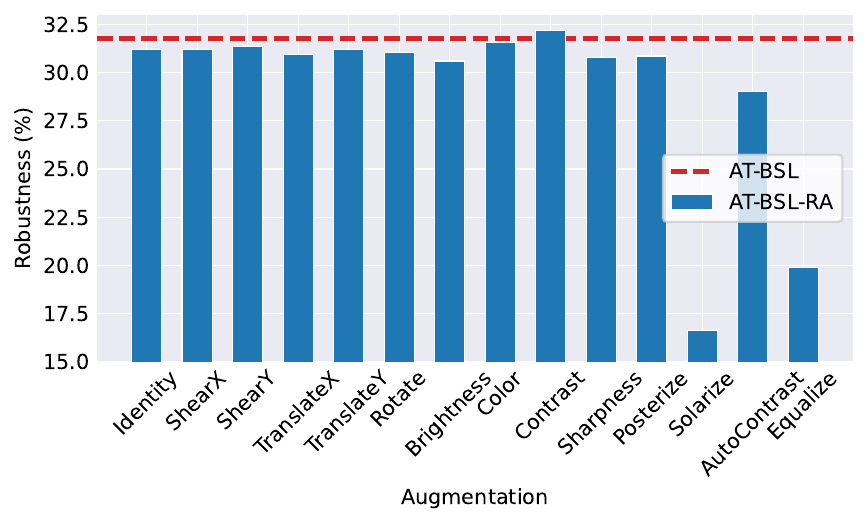} &
        \includegraphics[width=0.43\linewidth]{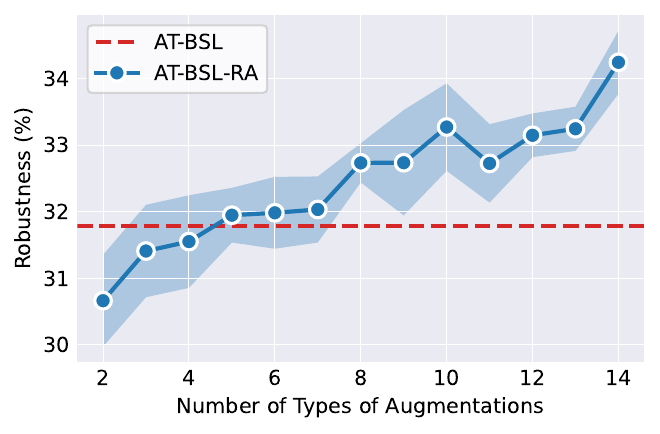}
        \\
        (a) & (b) 
    \end{tabular}
    \vspace{-0.1 in}
    \caption{The robustness under AA for AT-BSL with different augmentations using ResNet-18 on CIFAR-10-LT. (a) Change the augmentation space of RA~\cite{cubuk2020randaugment} to a single augmentation, and the horizontal axis represents the name of the single augmentation. (b) The horizontal axis represents the number of types of augmentations in the search space of RA.}
    \label{fig_ra_single}
    \vspace{-0.1 in}
\end{figure}

\noindent \textbf{Formulating Hypothesis.} We postulate that data augmentation improves robustness by increasing example diversity, thereby allowing models to learn richer representations. Taking RA~\cite{cubuk2020randaugment} as an illustrative example, for each training image, RA randomly selects a series of augmentations from a search space consisting of 14 augmentations, namely Identity, ShearX, ShearY, TranslateX, TranslateY, Rotate, Brightness, Color, Contrast, Sharpness, Posterize, Solarize, AutoContrast, and Equalize, to apply to the image. We initiate an ablation study on RA, testing the impact of each augmentation individually. Specifically, we narrow the search space of RA to a single augmentation, meaning RA is restricted to using only this one augmentation to augment all training examples. From Fig. \ref{fig_ra_single}(a), it can be observed that except for Contrast, none of the augmentations alone improve robustness; in fact, augmentations such as Solarize, AutoContrast, and Equalize significantly underperform compared to AT-BSL. We surmise that this is due to the limited example diversity provided by a single augmentation, thereby resulting in no substantial improvement in robustness.

\noindent \textbf{Validating Hypothesis.} 
Subsequently, we explore the impact of the number of types of augmentations on robustness. Specifically, for each trial, we randomly selected $n$ types of augmentations to constitute the search space of RA, with $n\in\{2,14\}$. Each experiment is repeated five times. As shown in Fig. \ref{fig_ra_single}(b), we reveal that robustness progressively improves with the addition of more augmentation methods in the search space of RA. This indicates that as the number of types of augmentations in the search space increases, the variety of augmentations available to examples also grows, leading to greater example diversity. Consequently, the representations learned by the model become more comprehensive, thereby improving robustness. This validates our hypothesis.

Moreover, to further substantiate our hypothesis, we conduct an ablation study on the three types of augmentations—Solarize, AutoContrast, and Equalize—which, when used individually, impair robustness. Specifically, we eliminate these three and employ the remaining 11 augmentations as the baseline: RA-11. We then incrementally add one to three of the negative augmentations, with the results outlined in Table \ref{tab_ra_ablation}. It is discovered that the more types of augmentations added, the more significant the improvement in robustness. Despite the negative impact of these three augmentations when used in isolation, their inclusion in the search space of RA still contributes to robustness improvement, further validating our hypothesis that data augmentation increases example diversity and thereby improves robustness.

\begin{table}[t]
\begin{center}
\caption{The clean accuracy and robustness under AA for AT-BSL with different augmentations using ResNet-18 on CIFAR-10-LT. The best results are \textbf{bolded}. RA-11 means only using the first 11 augmentations in the search space of RA. The lines below RA-11 indicate additional augmentations based on RA-11, and the last line uses the complete search space of RA. SO: Solarize; AC: AutoContrast; EQ: Equalize.}
\label{tab_ra_ablation}
\vspace{-0.1 in}
\setlength{\tabcolsep}{1.1mm}{\begin{tabular}{ccccccc}
\toprule[1.0pt]
Method   & Clean          & FGSM           & PGD            & CW             & LSA            & AA             \\ \midrule
RA-11        & 67.80          & 40.68          & 35.88          & 34.01          & 33.89          & 32.12          \\ \midrule
SO       & 67.60          & 41.43          & 37.04          & 34.52          & 34.05          & 32.76          \\
AC       & 68.57          & 41.20          & 36.60          & 34.24          & 34.07          & 32.51          \\
EQ       & 68.33          & 41.64          & 36.80          & 34.33          & 34.17          & 32.59          \\
SO+AC    & 68.43          & 42.10          & 37.23          & 34.62          & 34.37          & 33.02          \\
SO+EQ    & 68.53          & 41.89          & 37.42          & 35.07          & 34.83          & 33.49          \\
AC+EQ    & 68.36          & 41.88          & 37.42          & 34.91          & 34.49          & 33.15          \\
SO+AC+EQ & \textbf{70.86} & \textbf{43.06} & \textbf{37.94} & \textbf{36.24} & \textbf{36.04} & \textbf{34.24} \\ \bottomrule[1.0pt]
\end{tabular}}
\end{center}
\vspace{-0.25 in}
\end{table}

\section{Experiments}
\subsection{Settings}
\label{subsection_settings}
\noindent \textbf{Datasets.} Following ~\cite{wu2021adversarial}, we conduct experiments on CIFAR-10-LT and CIFAR-100-LT~\cite{krizhevsky2009learning}. Due to space constraints, partial results for CIFAR-100-LT are included in the appendix. In our main experiments, the imbalance ratio (IR) of CIFAR-10-LT is set to 50. Table \ref{tab_different_irs} also provides results for various IRs.

\noindent \textbf{Evaluation Metrics.} When assessing model robustness, the $l_{\infty}$ norm-bounded perturbation is $\epsilon=8 / 255$. The attacks carried out include the single-step attack FGSM~\cite{goodfellow2014explaining} and several iterative attacks, such as PGD~\cite{madry2018towards}, CW~\cite{carlini2017towards} and LSA~\cite{hitaj2021evaluating}, performed over 20 steps with a step size of $2 / 255$. We also employ AutoAttack (AA)~\cite{croce2020reliable}, considered the strongest attack so far. For all methods, the evaluations are based on both the best checkpoint (selected based on robustness under PGD-20) and the final checkpoint.

\noindent \textbf{Comparison Methods.} We consider adversarial training methods under long-tailed distributions: RoBal~\cite{wu2021adversarial} and REAT~\cite{li2023adversarial}, as well as defenses under balanced distributions: AT~\cite{madry2018towards}, TRADES~\cite{zhang2019theoretically}, MART~\cite{wang2019improving}, AWP~\cite{wu2020adversarial}, GAIRAT~\cite{zhang2020geometry}, and LAS-AT~\cite{jia2022adversarial}.

\noindent \textbf{Training Details.} 
We train the models using the Stochastic Gradient Descent (SGD) optimizer with an initial learning rate of 0.1, momentum of 0.9, and weight decay of 5e-4. We set the batch size to 128. We set the total number of training epochs to 100, and the learning rate is divided by 10 at the 75th and 90th epoch following~\cite{zhang2019theoretically}. During generating adversarial examples, we enforce a maximum perturbation of $8/255$ and a step size of $2/255$. The number of iterations for internal maximization is fixed at 10, denoting PGD-10, and the impact of PGD steps on robustness is investigated in Table \ref{tab_different_pgd_steps}. For all experiments related to AT-BSL, we adopt $\tau_b = 1$, and the results for different $\tau_b $ are provided in Fig. \ref{fig_different_taub}. Note that the AT-BSL presented in Tables \ref{tab_res_cifar10} and \ref{tab_wrn_cifar10} represents our own implementation, which differs in training parameters from RoBal~\cite{wu2021adversarial}. Detailed discussions regarding these discrepancies are provided in the appendix.

\subsection{Main Results}
As evident from Tables \ref{tab_res_cifar10} and \ref{tab_wrn_cifar10}, on CIFAR-10-LT, AT-BSL with data augmentation achieves the highest clean accuracy and adversarial robustness on both ResNet-18 and WideResNet-34-10. Note that on WideResNet-34-10, our method, AT-BSL-AuA, demonstrates a significant improvement of +6.66\% robustness under AA compared to the SOTA method RoBal. Moreover, in terms of robustness at the final checkpoint, our method significantly outperforms others, demonstrating that data augmentation mitigates robust overfitting.

\begin{table*}[t]
\begin{center}
\caption{The clean accuracy and robustness for various algorithms using ResNet-18 on CIFAR-10-LT. The best results are \textbf{bolded}.}
\label{tab_res_cifar10}
\vspace{-0.1 in}
\begin{tabular}{ccccccccccccc}
\toprule[1.0pt]
\multirow{2}{*}{Method} & \multicolumn{6}{c}{Best   Checkpoint}                                                               & \multicolumn{6}{c}{Last   Checkpoint}                                                               \\ \cmidrule(l){2-7} \cmidrule(l){8-13}
                        & Clean          & FGSM           & PGD            & CW             & LSA            & AA             & Clean          & FGSM           & PGD            & CW             & LSA            & AA             \\ \midrule
AT \cite{madry2018towards}                     & 49.35          & 30.09          & 27.30          & 26.93          & 27.08          & 25.76          & 52.91          & 29.29          & 25.15          & 25.58          & 27.13          & 24.23          \\
TRADES \cite{zhang2019theoretically}                  & 43.61          & 29.18          & 27.81          & 26.73          & 26.58          & 26.41          & 43.75          & 29.06          & 27.05          & 26.10          & 25.93          & 25.78          \\
MART \cite{wang2019improving}                    & 48.61          & 32.75          & 30.29          & 28.82          & 28.46          & 27.73          & 48.80          & 32.60          & 29.78          & 28.45          & 28.12          & 27.30          \\
AWP \cite{wu2020adversarial}                     & 49.29          & 33.78          & 31.20          & 30.53          & 30.36          & 29.53          & 47.75          & 32.77          & 30.83          & 30.01          & 29.68          & 29.12          \\
GAIRAT \cite{zhang2020geometry}                  & 50.83          & 30.20          & 27.46          & 21.65          & 21.23          & 20.41          & 50.66          & 28.44          & 25.60          & 19.68          & 19.22          & 18.26          \\
LAS-AT\cite{jia2022adversarial}                  & 52.81          & 33.35          & 30.32          & 29.57          & 29.15          & 28.53          & 53.50          & 33.14          & 30.09          & 29.13          & 28.84          & 28.30          \\ \midrule
RoBal \cite{wu2021adversarial}                   & 70.34          & 40.50          & 35.93          & 31.05          & 31.10          & 29.54          & 70.00          & 36.18          & 29.00          & 27.67          & 26.98          & 25.63          \\
REAT \cite{li2023adversarial}                    & 67.38          & 40.13          & 35.83          & 33.88          & 33.66          & 32.20          & 67.58          & 36.99          & 30.93          & 30.83          & 31.62          & 28.61          \\
AT-BSL                 & 68.89          & 40.08          & 35.27          & 33.47          & 33.46          & 31.78          & 67.63          & 35.20          & 28.65          & 28.91          & 31.35          & 26.97          \\
AT-BSL-RA              & \textbf{70.86} & \textbf{43.06} & \textbf{37.94} & \textbf{36.24} & \textbf{36.04} & \textbf{34.24} & \textbf{71.83} & \textbf{42.62} & \textbf{37.15} & \textbf{35.37} & \textbf{35.50} & \textbf{33.44} \\ \bottomrule[1.0pt]
\end{tabular}
\end{center}
\vspace{-0.15 in}
\end{table*}

\begin{table*}[t]
\begin{center}
\caption{The clean accuracy and robustness for various algorithms using WideResNet-34-10 on CIFAR-10-LT. The best results are \textbf{bolded}.}
\label{tab_wrn_cifar10}
\vspace{-0.1 in}
\begin{tabular}{ccccccccccccc}
\toprule[1.0pt]
\multirow{2}{*}{Method} & \multicolumn{6}{c}{Best   Checkpoint}                                                               & \multicolumn{6}{c}{Last   Checkpoint}                                                               \\ \cmidrule(l){2-7} \cmidrule(l){8-13} 
                        & Clean          & FGSM           & PGD            & CW             & LSA            & AA             & Clean          & FGSM           & PGD            & CW             & LSA            & AA             \\ \midrule
AT \cite{madry2018towards}                     & 59.21          & 31.88          & 27.88          & 28.19          & 29.81          & 27.07          & 58.25          & 29.77          & 25.29          & 25.71          & 29.83          & 24.94          \\
TRADES \cite{zhang2019theoretically}                  & 51.28          & 31.58          & 28.70          & 28.45          & 28.36          & 27.72          & 53.85          & 30.44          & 26.23          & 26.57          & 26.77          & 25.59          \\
MART \cite{wang2019improving}                    & 49.13          & 34.33          & 32.32          & 30.73          & 30.13          & 29.60          & 52.48          & 33.95          & 31.09          & 29.64          & 29.43          & 28.67          \\
AWP \cite{wu2020adversarial}                     & 50.91          & 34.28          & 31.85          & 31.23          & 31.01          & 30.06          & 48.65          & 33.21          & 31.07          & 30.33          & 30.14          & 29.40          \\
GAIRAT \cite{zhang2020geometry}                  & 59.89          & 33.47          & 30.40          & 26.69          & 26.71          & 25.38          & 56.37          & 29.41          & 27.25          & 23.94          & 23.95          & 23.15          \\
LAS-AT \cite{jia2022adversarial}                  & 57.52          & 33.66          & 29.86          & 29.60          & 29.44          & 28.84          & 58.19          & 32.98          & 28.89          & 28.75          & 28.58          & 27.90          \\ \midrule
RoBal \cite{wu2021adversarial}                   & 72.82          & 41.34          & 36.42          & 32.48          & 31.95          & 30.49          & 70.85          & 35.95          & 27.74          & 27.59          & 26.76          & 25.71          \\
REAT \cite{li2023adversarial}                    & 73.16          & 41.32          & 35.94          & 35.28          & 35.67          & 33.20          & 67.76          & 34.51          & 27.75          & 28.17          & 31.82          & 26.66          \\
AT-BSL                 & 73.19          & 41.84          & 35.60          & 34.86          & 35.99          & 32.80          & 65.95          & 33.29          & 27.23          & 27.87          & 31.00          & 26.45          \\
AT-BSL-AuA             & \textbf{75.17} & \textbf{46.18} & \textbf{40.84} & \textbf{38.82} & \textbf{39.23} & \textbf{37.15} & \textbf{77.27} & \textbf{44.73} & \textbf{38.06} & \textbf{37.14} & \textbf{39.05} & \textbf{35.11} \\ \bottomrule[1.0pt]
\end{tabular}
\end{center}
\vspace{-0.2 in}
\end{table*}

\begin{figure}[t]
    \centering
    \begin{tabular}{cc}
        \includegraphics[width=0.45\linewidth]{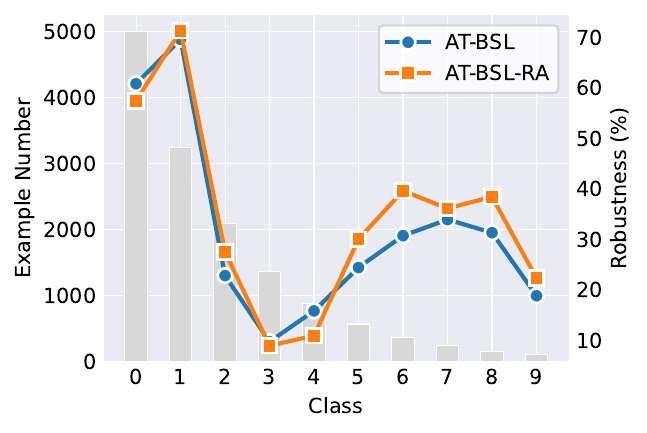} &
        \includegraphics[width=0.45\linewidth]{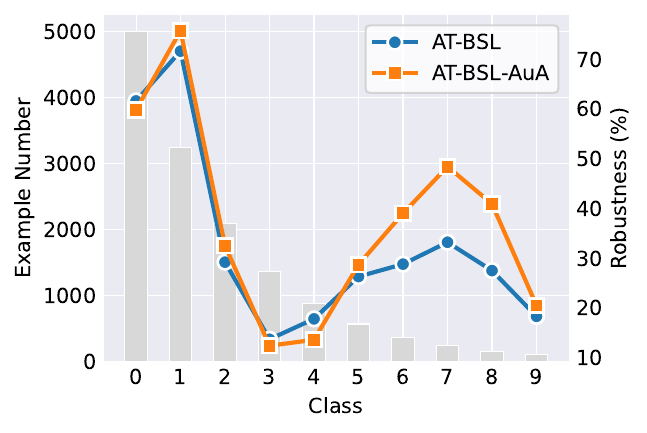} \\
        (a) & (b) \\
    \end{tabular}
    \vspace{-0.1 in}
    \caption{The class-wise example number and robustness under AA for various algorithms on CIFAR-10-LT at the best checkpoint. (a) ResNet-18; (b) WideResNet-34-10.}
    \label{fig_class_wise}
    \vspace{-0.25 in}
\end{figure}

We present the robustness of different methods across each class in Fig. \ref{fig_class_wise}. It is observable that, except for a few classes, our method improves robustness in almost every class, particularly in tail classes (5 to 9 classes) where the improvements are more pronounced. Furthermore, consistent with observations on balanced datasets~\cite{xu2021robust,ma2022tradeoff,wei2023cfa,yue2023revisiting}, there is a significant disparity in class-wise robustness. Class 3 remains the least robust despite its example numbers far exceeding that of subsequent classes, which may be attributable to the intrinsic properties of class 3~\cite{wu2021adversarial}.

\subsection{Futher Analysis}

\begin{figure}[t]
    \centering
    \begin{tabular}{cc}
        \includegraphics[width=0.45\linewidth]{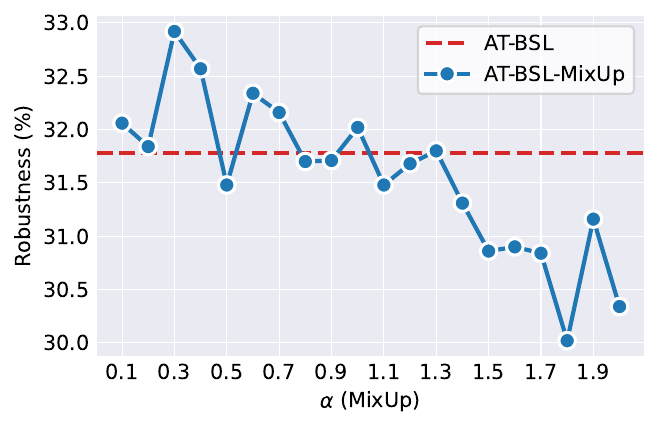} &
        \includegraphics[width=0.45\linewidth]{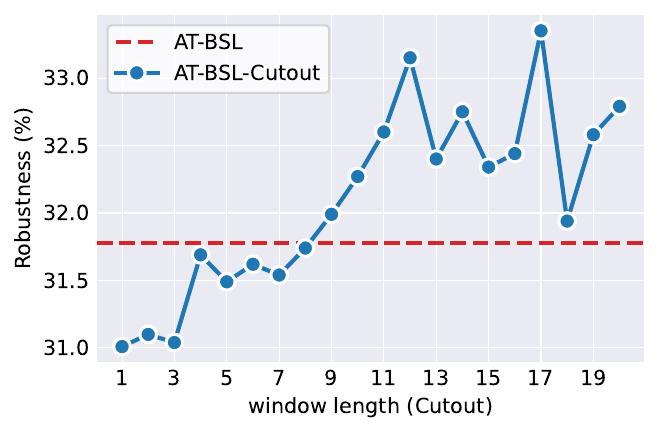} \\
        (a) & (b) \\
        \includegraphics[width=0.45\linewidth]{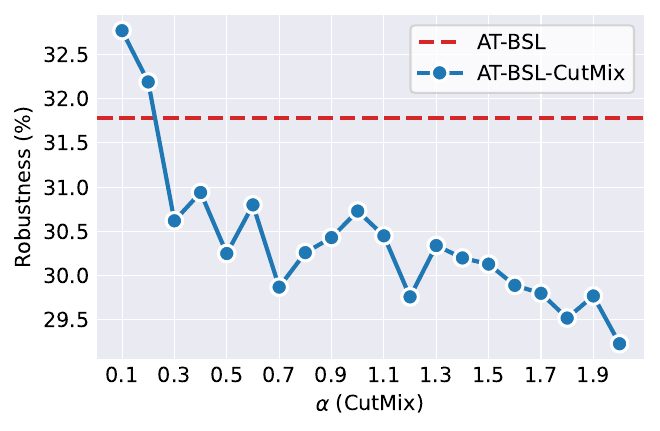} &
        \includegraphics[width=0.45\linewidth]{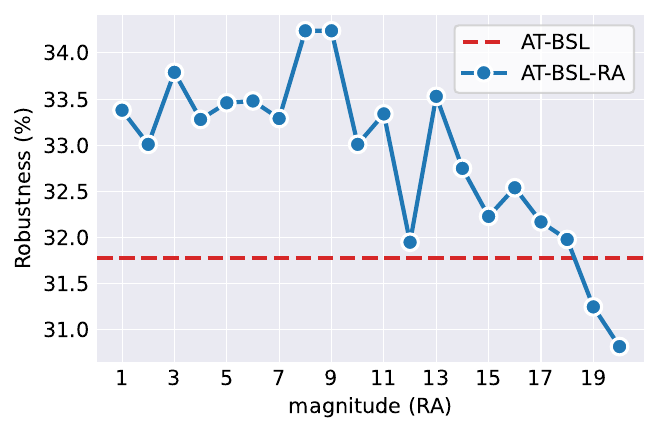} \\
        (c) & (d) 
    \end{tabular}
    \vspace{-0.1 in}
    \caption{The robustness under AA using ResNet-18 on CIFAR-10-LT as we vary (a) the mixing rate $\alpha$ for MixUp, (b) the window length for Cutout, (c) the mixing rate $\alpha$ for CutMix, and (d) the magnitude of transformations for RA.}
    \label{fig_aug_hyber}
\end{figure}

\noindent \textbf{Effect of Augmentation Strategies and Parameters.} We present in both Table \ref{tab_different_aug} and Fig. \ref{fig_aug_hyber} the impact of different augmentation strategies and parameters on robustness. Specifically, we conduct experiments using ResNet-18 on CIFAR-10-LT, comparing robustness at the best checkpoint. In addition, in Table \ref{tab_different_aug}, we use the best hyper-parameters: mixing rate $\alpha = 0.3$ for Mixup, window length 17 for Cutout, mixing rate $\alpha = 0.1$ for CutMix, and magnitude 8 for RA. As shown in Table \ref{tab_different_aug}, various augmentation strategies improve robustness compared to vanilla AT-BSL, with AuA and RA also achieving gains in clean accuracy. Fig. \ref{fig_aug_hyber} indicates that for MixUp and CutMix, smaller values of $\alpha$ yield better robustness; for Cutout, longer window lengths generally correlate with better robustness; for RA, a moderate magnitude of transformation improves robustness, peaking at $\text{magnitude}=8$, highlighting that excessive augmentation is not always beneficial.

\begin{table}[]
\begin{center}
\caption{The clean accuracy and robustness for AT-BSL with different augmentations using ResNet-18 on CIFAR-10-LT. The best results are \textbf{bolded}.}
\label{tab_different_aug}
\vspace{-0.1 in}
\setlength{\tabcolsep}{1.1mm}{
\begin{tabular}{ccccccc}
\toprule[1.0pt]
Method & Clean          & FGSM           & PGD            & CW             & LSA            & AA             \\ \midrule
Vanilla                       & 68.89          & 40.08          & 35.27          & 33.47          & 33.46          & 31.78          \\ \midrule
MixUp~\cite{zhang2018mixup}                       & 65.82          & 41.33          & \textbf{38.05} & 34.29          & 33.63          & 32.92          \\
Cutout~\cite{devries2017improved}                      & 65.12          & 40.25          & 36.68          & 34.81          & 34.51          & 33.35          \\
CutMix~\cite{yun2019cutmix}                      & 64.54          & 41.13          & 37.86          & 34.10          & 33.46          & 32.83          \\
AugMix~\cite{hendrycks2019augmix}                      & 67.12          & 40.31          & 35.95          & 34.19          & 34.02          & 32.51          \\
TA~\cite{muller2021trivialaugment}                          & 67.14          & 41.56          & 37.75          & 34.34          & 33.90          & 32.62          \\
AuA~\cite{cubuk2019autoaugment}                         & \textbf{71.63} & 42.69          & 37.78          & 35.60          & 35.47          & 33.69          \\
RA~\cite{cubuk2020randaugment}                          & 70.86          & \textbf{43.06} & 37.94          & \textbf{36.24} & \textbf{36.04} & \textbf{34.24}          \\ \bottomrule[1.0pt]
\end{tabular}}
\end{center}
\vspace{-0.25 in}
\end{table}

\begin{figure}[t]
    \centering
    \begin{tabular}{cc}
        \includegraphics[width=0.45\linewidth]{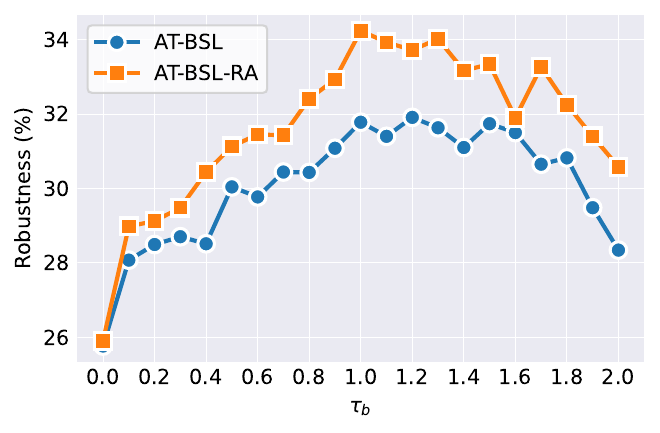} &
        \includegraphics[width=0.45\linewidth]{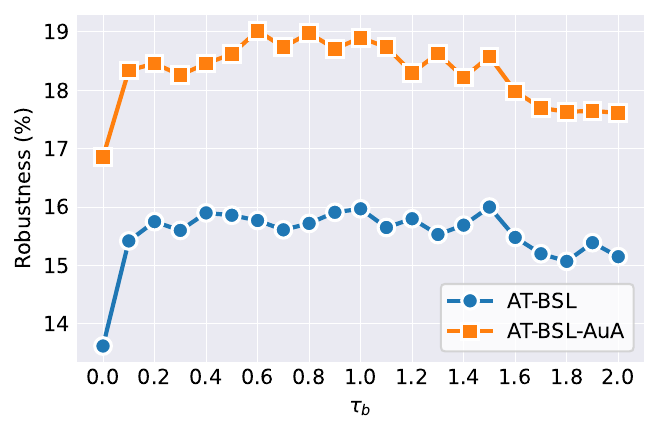} \\
        (a)& (b)
    \end{tabular}
    \vspace{-0.1 in}
    \caption{The robustness under AA for various algorithms with different $\tau_b$ using ResNet-18. (a): CIFAR-10-LT; (b): CIFAR-100-LT.}
    \label{fig_different_taub}
    \vspace{-0.1 in}
\end{figure}

\noindent \textbf{Effect of Hyperparameter $\tau_b$.} To investigate the sensitivity of AT-BSL to $\tau_b$, we evaluate the performance of AT-BSL under varying $\tau_b$ values. Specifically, we utilize ResNet-18 with $\tau_b$ ranging from 0 to 20. Note that at $\tau_b=0$, the bias $b_i=\tau_b \log \left(n_i\right)$ added by AT-BSL becomes zero, and Eq.\ref{eq_at_bsl} reverts to the vanilla CE loss, transforming AT-BSL into vanilla AT~\cite{madry2018towards}.  The results, depicted in Fig. \ref{fig_different_taub}, reveal that on CIFAR10-LT, AT-BSL is quite sensitive to $\tau_b$, achieving optimal robustness at $\tau_b=1$. Conversely, on CIFAR-100-LT, AT-BSL shows less sensitivity to $\tau_b$. Additionally, across tested datasets and $\tau_b$ values, AT-BSL with additional data augmentation consistently exhibits significantly higher robustness than vanilla AT-BSL, underscoring the substantial benefits of data augmentation in adversarial training under long-tailed distributions.

\noindent \textbf{Effect of Imbalance Ratio.} We further construct long-tailed datasets with varying IRs following the protocol of ~\cite{cui2019class,wu2021adversarial} to evaluate the performance of our method. Table \ref{tab_different_irs} illustrates that RA consistently improves the robustness of AT-BSL across various IR settings, further substantiating the finding that data augmentation can improve robustness.

\begin{table}[t]
\begin{center}
\caption{The clean accuracy and robustness for various algorithms using ResNet-18 on CIFAR-10-LT with different imbalance ratios. Better results are \textbf{bolded}.}
\label{tab_different_irs}
\vspace{-0.1 in}
\setlength{\tabcolsep}{0.7mm}{\begin{tabular}{cccccccc}
\toprule[1.0pt]
IR                   & Method    & Clean          & FGSM           & PGD            & CW             & LSA            & AA             \\ \midrule
\multirow{2}{*}{10}  & AT-BSL    & 73.29          & 47.33          & 42.04          & 40.77          & 41.05          & 39.12          \\
                     & AT-BSL-RA & \textbf{79.00} & \textbf{50.98} & \textbf{44.19} & \textbf{42.82} & \textbf{43.10} & \textbf{40.56} \\ \midrule
\multirow{2}{*}{20}  & AT-BSL    & 71.89          & 44.76          & 39.40          & 38.47          & 38.68          & 36.74          \\
                     & AT-BSL-RA & \textbf{75.84} & \textbf{47.62} & \textbf{41.68} & \textbf{39.92} & \textbf{39.82} & \textbf{37.78} \\ \midrule
\multirow{2}{*}{50}  & AT-BSL    & 68.89          & 40.08          & 35.27          & 33.47          & 33.46          & 31.78          \\
                     & AT-BSL-RA & \textbf{70.86} & \textbf{43.06} & \textbf{37.94} & \textbf{36.24} & \textbf{36.04} & \textbf{34.24} \\ \midrule
\multirow{2}{*}{100} & AT-BSL    & 62.03          & 35.06          & 30.95          & 29.41          & 29.56          & 28.01          \\
                     & AT-BSL-RA & \textbf{66.85} & \textbf{38.75} & \textbf{33.69} & \textbf{31.77} & \textbf{31.50} & \textbf{30.00} \\   \bottomrule[1.0pt]
\end{tabular}}
\end{center}
\vspace{-0.2 in}
\end{table}

\noindent \textbf{Effect of PGD Step Size.} To delve into the impact of PGD step size on robustness, we fine-tune the PGD step size from $2/255$ to $1/255$ and $0.5/255$, while also increasing the PGD steps from 10 to 20 and 40. As depicted in Table \ref{tab_different_pgd_step_size}, it is evident that RA consistently improves the robustness of AT-BSL regardless of the PGD step size. However, we also note a decrease in robustness when compared to the baseline robustness at a PGD step size of $2/255$.

\begin{table}[t]
\begin{center}
\caption{The clean accuracy and robustness for various algorithms using ResNet-18 on CIFAR-10-LT training with different PGD step sizes. Better results are \textbf{bolded}.}
\label{tab_different_pgd_step_size}
\vspace{-0.1 in}
\setlength{\tabcolsep}{0.6mm}{\begin{tabular}{cccccccc}
\toprule[1.0pt]
Size                 & Method    & Clean          & FGSM           & PGD            & CW             & LSA            & AA             \\ \midrule
\multirow{2}{*}{0.5} & AT-BSL    & 68.57          & 39.65          & 35.10          & 32.92          & 32.97          & 31.28          \\
                     & AT-BSL-RA & \textbf{68.68} & \textbf{41.97} & \textbf{37.60} & \textbf{34.81} & \textbf{34.36} & \textbf{33.26} \\ \midrule
\multirow{2}{*}{1}   & AT-BSL    & 68.63          & 39.98          & 35.09          & 33.02          & 33.00          & 31.18          \\
                     & AT-BSL-RA & \textbf{68.93} & \textbf{42.71} & \textbf{37.85} & \textbf{35.30} & \textbf{34.79} & \textbf{33.51} \\ \midrule
\multirow{2}{*}{2}   & AT-BSL    & 68.89          & 40.08          & 35.27          & 33.47          & 33.46          & 31.78          \\
                     & AT-BSL-RA & \textbf{70.86} & \textbf{43.06} & \textbf{37.94} & \textbf{36.24} & \textbf{36.04} & \textbf{34.24} \\ \bottomrule[1.0pt]
\end{tabular}}
\end{center}
\vspace{-0.25 in}
\end{table}

\section{Conclusion}

In this paper, we first dissect the components of RoBal, identifying BSL as a critical component. We then address the issue of robust overfitting in adversarial training under long-tailed distributions and attempt to mitigate it using data augmentation. Surprisingly, we find that data augmentation not only mitigates robust overfitting but also significantly improves robustness. We hypothesize that the improved robustness is due to increased example diversity brought about by data augmentation, and we validate this hypothesis through experiments. Finally, we conduct extensive experiments with different data augmentation strategies, model architectures, and datasets, affirming the generalizability of our findings. Our findings advance adversarial training a step further towards real-world scenarios.

\section*{Acknowledgements}
This work was partially supported by the NSFC under Grants U20B2049, U21B2018, and 62302344, and the Fundamental Research Funds for the Central Universities, 2042023kf0122.
\clearpage

{
    \small
    \bibliographystyle{ieeenat_fullname}
    \bibliography{main}
}


\clearpage
\setcounter{page}{1}
\maketitlesupplementary
\appendix
\section{Implementation Details of Experiments}
\subsection{Details of Table \ref{tab_robal_ablation}}

All parameters in the experiments presented in Table \ref{tab_robal_ablation} are consistent with those used in RoBal~\cite{wu2021adversarial}. Specifically, the initial learning rate is set at 0.1, with a decay factor of 10 applied at the 60th and 75th epochs, for a total training duration of 80 epochs. An SGD momentum optimizer is employed with a weight decay of $2\times10^{-4}$. The batch size is maintained at 64. For adversarial training, we adopt a maximum perturbation of $8/255$ and a step size of $2/255$, with the number of iterations for internal maximization set at 5, corresponding to PGD-5. For CIFAR-10-LT, we utilize $m_0=0.1$, $s=10$, $\tau_b=1.5$, and $\tau_m=0.3$; for CIFAR-100-LT, the parameters are set as $m_0=0.3$, $s=10$, $\tau_b=1.5$, and $\tau_m=0.3$. The specific hyperparameters for each experiment are detailed in Table \ref{tab_robal_ablation_hyperparameters}.

\begin{table*}[t]
\begin{center}
\caption{The specific hyperparameters for each experiment following the integration of components from RoBal~\cite{wu2021adversarial} into AT~\cite{madry2018towards}. Cos: Cosine Classifier; BSL: Balanced Softmax Loss~\cite{ren2020balanced}; CM: Class-aware Margin~\cite{wu2021adversarial}; TRADES: TRADES Regularization~\cite{zhang2019theoretically}.}
\label{tab_robal_ablation_hyperparameters}
\vspace{-0.1 in}
\begin{tabular}{ccccccccccccc}
\toprule[1.0pt]
\multirow{2}{*}{Method} & \multicolumn{4}{c}{Components} & \multicolumn{4}{c}{CIFAR-10-LT} & \multicolumn{4}{c}{CIFAR-100-LT} \\ \cmidrule(l){2-5} \cmidrule(l){6-9} \cmidrule(l){10-13} 
                        & Cos   & BSL   & CM   & TRADES  & $m_0$    & $s$   & $\tau_b$  & $\tau_m$  & $m_0$    & $s$    & $\tau_b$  & $\tau_m$  \\ \midrule
AT~\cite{madry2018towards}                      &       &       &      &         & 0     & 1   & 0       & 0       & 0     & 1    & 0       & 0       \\
AT-BSL                  &       & \checkmark     &      &         & 0     & 1   & 1       & 0       & 0     & 1    & 1       & 0       \\
AT-BSL-Cos              & \checkmark     & \checkmark     &      &         & 0.1   & 10  & 1       & 0       & 0.3   & 10   & 1       & 0       \\
AT-BSL-Cos-TRADES       & \checkmark     & \checkmark     &      & \checkmark       & 0.1   & 10  & 1.5     & 0       & 0.3   & 10   & 1.5     & 0       \\
RoBal~\cite{wu2021adversarial}                 & \checkmark     & \checkmark     & \checkmark    & \checkmark       & 0.1   & 10  & 1.5     & 0.3     & 0.3   & 10   & 1.5     & 0.3     \\ \bottomrule[1.0pt]
\end{tabular}
\end{center}
\vspace{-0.1 in}
\end{table*}

\subsection{Details of Data Augmentaions}

Data augmentation techniques such as MixUp~\cite{zhang2018mixup}, Cutout~\cite{devries2017improved}, CutMix~\cite{yun2019cutmix}, Augmix~\cite{hendrycks2019augmix}, AutoAugment (AuA)~\cite{cubuk2019autoaugment}, RandAugment (RA)~\cite{cubuk2020randaugment}, and TrivialAugmen (TA)~\cite{muller2021trivialaugment} are employed utilizing the implementations provided in torchvision 0.16.0\footnote{\href{https://github.com/pytorch/pytorch}{https://github.com/pytorch/pytorch}}. Regarding the integration of data augmentation into the adversarial training pipeline, we follow the approach outlined in ~\cite{rebuffi2021data}, whereby data augmentation precedes the generation of adversarial examples through adversarial attacks. It is observed that reversing this order, i.e., performing data augmentation after adversarial attacks, leads to the disruption of adversarial perturbations, significantly diminishing the effectiveness of the adversarial attacks.

\subsection{Code References}
For the defense methods compared in our paper, we utilize the official code releases, including AT~\cite{madry2018towards}\footnote{\href{https://github.com/MadryLab/cifar10_challenge}{https://github.com/MadryLab/cifar10\_challenge}}, TRADES~\cite{zhang2019theoretically}\footnote{\href{https://github.com/yaodongyu/TRADES}{https://github.com/yaodongyu/TRADES}}, MART~\cite{wang2019improving}\footnote{\href{https://github.com/YisenWang/MART}{https://github.com/YisenWang/MART}}, AWP~\cite{wu2020adversarial}\footnote{\href{https://github.com/csdongxian/AWP}{https://github.com/csdongxian/AWP}}, GAIRAT~\cite{zhang2020geometry}\footnote{\href{https://github.com/zjfheart/Geometry-aware-Instance-reweighted-Adversarial-Training}{https://github.com/zjfheart/Geometry-aware-Instance-reweighted-Adversarial-Training}}, LAS-AT~\cite{jia2022adversarial}\footnote{\href{https://github.com/jiaxiaojunqaq/las-at}{https://github.com/jiaxiaojunqaq/las-at}}, RoBal~\cite{wu2021adversarial}\footnote{\href{https://github.com/wutong16/Adversarial_Long-Tail}{https://github.com/wutong16/Adversarial\_Long-Tail}}, and REAT~\cite{li2023adversarial}\footnote{\href{https://github.com/GuanlinLee/REAT}{https://github.com/GuanlinLee/REAT}}. Regarding the attacks used for evaluation, we implement them by referring to several official code repositories and the original papers, encompassing FGSM~\cite{goodfellow2014explaining}, PGD~\cite{madry2018towards}, CW~\cite{carlini2017towards}, and AutoAttack~\cite{croce2020reliable}\footnote{\href{https://github.com/fra31/auto-attack}{https://github.com/fra31/auto-attack}}.

\section{Extensive Experiments}

\subsection{More Ablation Studies of RoBal}
In addition to the experiments conducted with ResNet-18 and CIFAR-10-LT as presented in Table \ref{tab_robal_ablation}, we extend our ablation studies to include WideResNet-34-10 and CIFAR-100-LT, as illustrated in Tables \ref{tab_robal_ablation_wrn_cifar10}, \ref{tab_robal_ablation_res_cifar100}, and \ref{tab_robal_ablation_wrn_cifar100}. The data acquired from these additional experiments align with the conclusions drawn from Table \ref{tab_robal_ablation}, demonstrating that the AT-BSL alone achieves comparable performance in terms of clean accuracy and robustness to the complete RoBal framework. Moreover, a significant advantage is observed regarding training time and memory consumption.

\begin{table*}[ht]
\begin{center}
\caption{The clean accuracy, robustness, time (average per epoch), and memory (GPU) using WideResNet-34-10~\cite{zagoruyko2016wide} on CIFAR-10-LT following the integration of components from RoBal~\cite{wu2021adversarial} into AT~\cite{madry2018towards}. The best results are \textbf{bolded}. The second best results are \uline{underlined}. Cos: Cosine Classifier; BSL: Balanced Softmax Loss~\cite{ren2020balanced}; CM: Class-aware Margin~\cite{wu2021adversarial}; TRADES: TRADES Regularization~\cite{zhang2019theoretically}.}
\label{tab_robal_ablation_wrn_cifar10}
\vspace{-0.1 in}
\setlength{\tabcolsep}{1.0mm}{\begin{tabular}{ccccccccccccc}
\toprule[1.0pt]
\multirow{2}{*}{Method} & \multicolumn{4}{c}{Components} & \multicolumn{6}{c}{Accuracy}                                                                        & \multicolumn{2}{c}{Efficiency} \\ \cmidrule(l){2-5} \cmidrule(l){6-11} \cmidrule(l){12-13} 
                        & Cos   & BSL   & CM   & TRADES  & Clean          & FGSM           & PGD             & CW             & LSA            & AA             & Time (s)         & Memory (MiB)         \\ \midrule
AT~\cite{madry2018towards}                      &       &       &      &         & 60.86          & 33.22          & 28.79          & 29.24          & 31.27          & 27.66          & \textbf{160.01} & \textbf{2574} \\
AT-BSL                  &       & \checkmark     &      &         & \uline{73.84}    & 39.13          & 32.02          & \uline{32.29}    & \textbf{34.98} & \uline{30.21}    & \uline{162.01}    & \textbf{2574} \\
AT-BSL-Cos              & \checkmark     & \checkmark     &      &         & \textbf{74.69} & 40.86          & 34.77          & 31.14          & 30.50          & 29.22          & 163.71          & \textbf{2574} \\
AT-BSL-Cos-TRADES       & \checkmark     & \checkmark     &      & \checkmark       & 73.34          & \uline{41.28}    & \textbf{36.49} & 31.79          & 31.55          & 30.05          & 302.62          & \uline{6932}    \\
RoBal~\cite{wu2021adversarial}                 & \checkmark     & \checkmark     & \checkmark    & \checkmark       & 72.82          & \textbf{41.34} & \uline{36.42}    & \textbf{32.48} & \uline{31.95}    & \textbf{30.49} & 309.09          & \uline{6932}     \\ \bottomrule[1.0pt]
\end{tabular}}
\end{center}
\end{table*}

\begin{table*}[ht]
\begin{center}
\caption{The clean accuracy, robustness, time (average per epoch) and memory (GPU) using ResNet-18~\cite{he2016deep} on CIFAR-100-LT following the integration of components from RoBal~\cite{wu2021adversarial} into AT~\cite{madry2018towards}. The best results are \textbf{bolded}. The second best results are \uline{underlined}. Cos: Cosine Classifier; BSL: Balanced Softmax Loss~\cite{ren2020balanced}; CM: Class-aware Margin~\cite{wu2021adversarial}; TRADES: TRADES Regularization~\cite{zhang2019theoretically}.}
\label{tab_robal_ablation_res_cifar100}
\vspace{-0.1 in}
\setlength{\tabcolsep}{1.0mm}{\begin{tabular}{ccccccccccccc}
\toprule[1.0pt]
\multirow{2}{*}{Method} & \multicolumn{4}{c}{Components} & \multicolumn{6}{c}{Accuracy}                                                                        & \multicolumn{2}{c}{Efficiency} \\ \cmidrule(l){2-5} \cmidrule(l){6-11} \cmidrule(l){12-13} 
                        & Cos   & BSL   & CM   & TRADES  & Clean          & FGSM           & PGD             & CW             & LSA            & AA             & Time (s)           & Memory (MiB)      \\ \midrule
AT~\cite{madry2018towards}                      &       &       &      &         & 44.32          & 18.81          & 15.11          & 15.36          & 17.85          & 13.91          & \uline{43.25}     & \textbf{946} \\
AT-BSL                  &       & \checkmark     &      &         & \uline{45.78}    & \uline{21.58}    & \textbf{18.96} & \uline{17.78}    & \uline{18.48}    & \uline{16.35}    & \textbf{41.99}  & \textbf{946} \\
AT-BSL-Cos              & \checkmark     & \checkmark     &      &         & 41.83          & 17.95          & 14.69          & 14.22          & 14.87          & 13.14          & 43.86           & \textbf{946} \\
AT-BSL-Cos-TRADES       & \checkmark     & \checkmark     &      & \checkmark       & 37.50          & 16.92          & 14.05          & 13.98          & 14.52          & 12.87          & 72.34           & \uline{1724}   \\
RoBal~\cite{wu2021adversarial}                 & \checkmark     & \checkmark     & \checkmark    & \checkmark       & \textbf{45.93} & \textbf{21.35} & \uline{17.40}    & \textbf{17.80} & \textbf{19.14} & \textbf{16.42} & 72.93           & \uline{1724}   \\ \bottomrule[1.0pt]
\end{tabular}}
\end{center}
\vspace{-0.15 in}
\end{table*}

\begin{table*}[ht]
\begin{center}
\caption{The clean accuracy, robustness, time (average per epoch), and memory (GPU) using WideResNet-34-10~\cite{zagoruyko2016wide} on CIFAR-100-LT following the integration of components from RoBal~\cite{wu2021adversarial} into AT~\cite{madry2018towards}. The best results are \textbf{bolded}. The second best results are \uline{underlined}. Cos: Cosine Classifier; BSL: Balanced Softmax Loss~\cite{ren2020balanced}; CM: Class-aware Margin~\cite{wu2021adversarial}; TRADES: TRADES Regularization~\cite{zhang2019theoretically}.}
\label{tab_robal_ablation_wrn_cifar100}
\vspace{-0.1 in}
\setlength{\tabcolsep}{1.0mm}{\begin{tabular}{ccccccccccccc}
\toprule[1.0pt]
\multirow{2}{*}{Method} & \multicolumn{4}{c}{Components} & \multicolumn{6}{c}{Accuracy}                                                                        & \multicolumn{2}{c}{Efficiency} \\ \cmidrule(l){2-5} \cmidrule(l){6-11} \cmidrule(l){12-13} 
                        & Cos   & BSL   & CM   & TRADES  & Clean          & FGSM           & PGD             & CW             & LSA            & AA             & Time (s)         & Memory (MiB)         \\ \midrule
AT~\cite{madry2018towards}                      &       &       &      &         & 48.87          & 21.14          & 17.20          & 17.61          & \uline{21.23}    & 16.27          & \textbf{319.33} & \textbf{2574} \\
AT-BSL                  &       & \checkmark     &      &         & \uline{49.68}    & \textbf{23.08} & \textbf{19.81} & \textbf{19.47} & 21.19          & \uline{17.84}    & \uline{323.66}    & \textbf{2574} \\
AT-BSL-Cos              & \checkmark     & \checkmark     &      &         & 48.29          & 20.25          & 16.34          & 16.43          & 17.90          & 15.09          & 327.17          & \textbf{2574} \\
AT-BSL-Cos-TRADES       & \checkmark     & \checkmark     &      & \checkmark       & 44.37          & 18.94          & 15.48          & 15.70          & 17.02          & 14.43          & 603.99          & \uline{6936}    \\
RoBal~\cite{wu2021adversarial}                 & \checkmark     & \checkmark     & \checkmark    & \checkmark       & \textbf{50.08} & \uline{23.04}    & \uline{18.84}    & \uline{19.30}    & \textbf{21.87} & \textbf{17.90} & 617.73          & \uline{6936}     \\ \bottomrule[1.0pt]
\end{tabular}}
\end{center}
\vspace{-0.15 in}
\end{table*}

\subsection{Experiments on CIFAR-100-LT}
Tables \ref{tab_res_cifar100} and \ref{tab_wrn_cifar100} reveal that on CIFAR-100-LT, AT-BSL with data augmentation achieves the highest clean accuracy and adversarial robustness on both ResNet-18 and WideResNet-34-10. Compared to the improvement observed on CIFAR-10-LT, the improvements on CIFAR-100-LT are less pronounced, likely due to CIFAR-100-LT's more significant number of classes and fewer examples per class, making advancements more challenging.

\begin{figure}[t]
    \centering
    \begin{tabular}{cc}
        \includegraphics[width=0.45\linewidth]{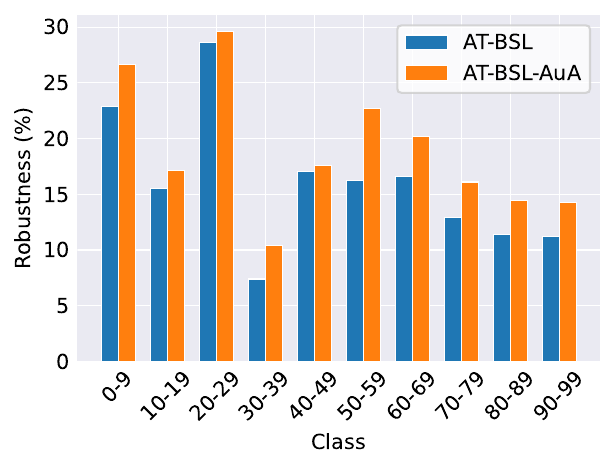} &
        \includegraphics[width=0.45\linewidth]{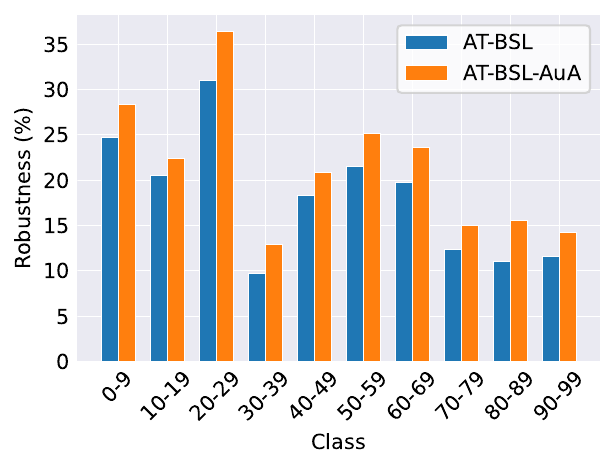} \\
        (a) & (b) \\
    \end{tabular}
    \vspace{-0.1 in}
    \caption{The class-wise robustness under AA for various algorithms on CIFAR-100-LT at the best checkpoint. (a) ResNet-18; (b) WideResNet-34-10.}
    \label{fig_class_wise_cifar100}          
    \vspace{-0.2 in}
\end{figure}

In Fig. \ref{fig_class_wise_cifar100}, we illustrate the robustness of different methods across each class. Given the extremely low robustness in most classes on CIFAR-100-LT and the presence of only 50 images per class in the test set, we report the average values for every 10 classes. Notably, AuA universally improves the robustness across all class groups.
\begin{table*}[t]
\begin{center}
\caption{The clean accuracy and robustness for various algorithms using ResNet-18 on CIFAR-100-LT. The best results are \textbf{bolded}.}
\label{tab_res_cifar100}
\vspace{-0.1 in}
\setlength{\tabcolsep}{2.0mm}{\begin{tabular}{ccccccccccccc}
\toprule[1.0pt]
\multirow{2}{*}{Method} & \multicolumn{6}{c}{Best   Checkpoint}                                                               & \multicolumn{6}{c}{Last   Checkpoint}                                                               \\ \cmidrule(l){2-7} \cmidrule(l){8-13}
                        & Clean          & FGSM           & PGD            & CW             & LSA            & AA             & Clean          & FGSM           & PGD            & CW             & LSA            & AA             \\ \midrule
AT \cite{madry2018towards}                     & 41.20          & 17.42          & 14.59          & 14.51          & 16.49          & 13.62          & 41.44          & 17.21          & 13.89          & 14.17          & 16.40          & 13.10          \\
TRADES \cite{zhang2019theoretically}                  & 38.12          & 19.60          & 17.89          & 15.96          & 15.91          & 15.59          & 38.71          & 19.43          & 17.27          & 15.83          & 15.87          & 15.28          \\
MART \cite{wang2019improving}                    & 38.46          & 23.04          & 21.36          & 18.59          & 18.36          & 17.51          & 39.58          & 22.38          & 20.51          & 18.40          & 18.42          & 17.27          \\
AWP \cite{wu2020adversarial}                     & 41.53          & 23.47          & 21.79          & 19.68          & 19.73          & 18.61         & 43.57          & 22.91          & 20.72          & 19.11          & 19.30          & 17.82          \\
GAIRAT \cite{zhang2020geometry}                  & 38.99          & 19.73          & 18.05          & 16.59          & 16.80          & 15.61          & 39.70          & 14.66          & 11.87          & 11.57          & 12.28          & 10.48          \\
LAS-AT \cite{jia2022adversarial}                  & 44.33          & 22.02          & 19.59          & 17.18          & 17.11          & 16.15          & 44.70          & 22.11          & 19.23          & 16.93          & 17.03          & 15.77          \\ \midrule
RoBal \cite{wu2021adversarial}                   & 45.93          & 21.35          & 17.40          & 17.80          & 19.14          & 16.42          & 45.78          & 19.97          & 15.37          & 15.75          & 18.67          & 14.51          \\
REAT \cite{li2023adversarial}                    & 46.28          & 21.55          & 18.85          & 18.07          & 18.95          & 16.54          & 45.99          & 19.62          & 16.29          & 16.04          & 18.22          & 14.79          \\
AT-BSL                 & 45.59          & 21.14          & 18.05          & 17.34          & 18.14          & 15.97          & 45.35          & 18.96          & 15.52          & 15.59          & 17.78          & 14.41          \\
AT-BSL-AuA             & \textbf{48.39} & \textbf{25.81} & \textbf{22.96} & \textbf{20.73} & \textbf{21.30} & \textbf{18.90}          & \textbf{50.66} & \textbf{25.89} & \textbf{22.43} & \textbf{20.62} & \textbf{21.43} & \textbf{18.79} \\ \bottomrule[1.0pt]
\end{tabular}}
\end{center}
\vspace{-0.1 in}
\end{table*}

\begin{table*}[t]
\begin{center}
\caption{The clean accuracy and robustness for various algorithms using WideResNet-34-10 on CIFAR-100-LT. The best results are \textbf{bolded}.}
\label{tab_wrn_cifar100}
\vspace{-0.1 in}
\setlength{\tabcolsep}{2.0mm}{\begin{tabular}{ccccccccccccc}
\toprule[1.0pt]
\multirow{2}{*}{Method} & \multicolumn{6}{c}{Best   Checkpoint}                                                               & \multicolumn{6}{c}{Last   Checkpoint}                                                               \\ \cmidrule(l){2-7} \cmidrule(l){8-13}
                        & Clean          & FGSM           & PGD            & CW             & LSA            & AA             & Clean          & FGSM           & PGD            & CW             & LSA            & AA             \\ \midrule
AT \cite{madry2018towards}                     & 45.18          & 19.25          & 16.36          & 16.43          & 19.00          & 15.60          & 44.86          & 19.01          & 15.65          & 15.89          & 19.12          & 15.08          \\
TRADES \cite{zhang2019theoretically}                  & 41.71          & 21.91          & 19.85          & 18.46          & 18.39          & 17.91          & 43.22          & 20.28          & 17.46          & 17.34          & 17.56          & 16.69          \\
MART \cite{wang2019improving}                    & 41.32          & 25.01          & 23.27          & 20.89          & 20.77          & 19.98          & 43.67          & 22.84          & 19.88          & 18.80          & 19.45          & 17.77          \\
AWP \cite{wu2020adversarial}                     & 45.66          & 25.89          & 23.88          & 21.87          & 22.10          & 20.56          & 48.18          & 24.75          & 21.81          & 20.30          & 21.19          & 18.67          \\
GAIRAT \cite{zhang2020geometry}                  & 36.41          & 18.87          & 17.31          & 16.07          & 16.13          & 14.77          & 45.11          & 19.49          & 16.31          & 15.85          & 16.71          & 14.75          \\
LAS-AT\cite{jia2022adversarial}                  & 45.86          & 23.30          & 20.02          & 18.67          & 18.79          & 17.35          & 46.54          & 22.84          & 19.65          & 18.18          & 18.38          & 17.01          \\ \midrule
RoBal \cite{wu2021adversarial}                   & 50.08          & 23.04          & 18.84          & 19.30          & 21.87          & 17.90          & 46.34          & 19.99          & 15.17          & 15.87          & 20.06          & 14.77          \\
REAT \cite{li2023adversarial}                    & 50.29          & 23.99          & 20.82          & 20.25          & 21.93          & 18.65          & 49.22          & 20.89          & 16.57          & 17.08          & 20.89          & 15.49          \\
AT-BSL                 & 50.04          & 23.37          & 19.66          & 19.60          & 21.66          & 18.04          & 48.56          & 20.88          & 16.83          & 17.09          & 20.13          & 15.76          \\
AT-BSL-AuA             & \textbf{53.08} & \textbf{28.55} & \textbf{25.40} & \textbf{23.39} & \textbf{24.48} & \textbf{21.43} & \textbf{55.55} & \textbf{26.74} & \textbf{22.18} & \textbf{21.88} & \textbf{24.28} & \textbf{19.68} \\ \bottomrule[1.0pt]
\end{tabular}}
\end{center}
\vspace{-0.2 in}
\end{table*}

\subsection{Experiments on Tiny-ImageNet-LT}

To see if BSL and data augmentation are as important for higher resolution datasets as they are for low resolution datasets (such as CIFAR-10-LT and CIFAR-100-LT), we conduct experiments on Tiny-ImageNet~\cite{le2015tiny}. Firstly, Tiny-ImageNet is a dataset consisting of 200 classes, with images sized 64*64 pixels, making it four times the resolution of CIFAR-10/100. We derive Tiny-ImageNet-LT using an IR of 0.1 from Tiny-ImageNet. Following ~\cite{avmixup,jia2022adversarial}, we employ the PreActResNet-18 model~\cite{preactresnet18}. Apart from the model, the experimental setup for Tiny-ImageNet-LT remains largely similar to that of CIFAR-10-LT. As observed from the Table \ref{tab_tiny}, both BSL and data augmentation prove to be crucial for Tiny-ImageNet-LT.

\begin{table}[t]
\begin{center}
\caption{The robustness for various algorithms with different training epochs using PreActResNet-18 on Tiny-ImageNet-LT at the best checkpoint. Better results are \textbf{bolded}.}
\label{tab_tiny}
\setlength{\tabcolsep}{1.1mm}{\begin{tabular}{ccccccc}
\toprule[1.0pt]
Method    & Clean          & FGSM           & PGD            & CW             & LSA            & AA             \\ \midrule
AT        & 36.30          & 16.58          & 14.52          & 12.65          & 13.16          & 11.37          \\ \midrule
RoBal     & 36.27          & 13.66          & 10.98          & 10.18          & 9.84           & 8.98           \\
AT-BSL    & 38.83          & 17.47          & 15.34          & 13.35          & 14.08          & 11.83          \\
AT-BSL-RA & \textbf{39.00} & \textbf{18.82} & \textbf{16.94} & \textbf{14.26} & \textbf{14.60} & \textbf{12.73} \\ \bottomrule[1.0pt]
\end{tabular}}
\end{center}
\vspace{-0.2 in}
\end{table}

\subsection{Different PGD Steps}
To investigate the impact of PGD steps on robustness, we assess the robustness achieved using different PGD steps following ~\cite{wu2021adversarial}. Table \ref{tab_different_pgd_steps} indicates that RA consistently improves the robustness of AT-BSL regardless of PGD steps, and the clean accuracy also experiences improvement. Moreover, there is a trade-off between clean accuracy and robustness: as the PGD step increases, clean accuracy decreases while robustness improves. The optimal trade-off is attained at PGD-10. Hence, we employ PGD-10 in our experiments involving AT-BSL.

\begin{table}[t]
\begin{center}
\caption{The clean accuracy and robustness for various algorithms using ResNet-18 on CIFAR-10-LT training with different PGD steps. Better results are \textbf{bolded}.}
\label{tab_different_pgd_steps}
\vspace{-0.1 in}
\setlength{\tabcolsep}{0.5mm}{\begin{tabular}{cccccccc}
\toprule[1.0pt]
Steps               & Method    & Clean          & FGSM           & PGD            & CW             & LSA            & AA             \\ \midrule
\multirow{2}{*}{1}  & AT-BSL    & 77.15          & 23.15          & 12.05          & 13.06          & 24.80          & 11.27          \\
                    & AT-BSL-RA & \textbf{82.16} & \textbf{28.28} & \textbf{14.25} & \textbf{15.34} & \textbf{26.30} & \textbf{13.21} \\ \midrule
\multirow{2}{*}{3}  & AT-BSL    & 72.37          & 36.61          & 28.95          & 28.79          & 30.23          & 26.64          \\
                    & AT-BSL-RA & \textbf{74.20} & \textbf{40.39} & \textbf{32.63} & \textbf{32.25} & \textbf{33.31} & \textbf{29.79} \\ \midrule
\multirow{2}{*}{5}  & AT-BSL    & 68.62          & 39.11          & 33.67          & 32.47          & 32.62          & 30.49          \\
                    & AT-BSL-RA & \textbf{69.39} & \textbf{41.86} & \textbf{36.81} & \textbf{34.33} & \textbf{33.89} & \textbf{32.62} \\ \midrule
\multirow{2}{*}{7}  & AT-BSL    & 68.28          & 39.55          & 34.62          & 32.94          & 32.68          & 31.16          \\
                    & AT-BSL-RA & \textbf{68.79} & \textbf{42.45} & \textbf{37.78} & \textbf{35.31} & \textbf{34.98} & \textbf{33.57} \\ \midrule
\multirow{2}{*}{10} & AT-BSL    & 68.89          & 40.08          & 35.27          & 33.47          & 33.46          & 31.78          \\
                    & AT-BSL-RA & \textbf{70.86} & \textbf{43.06} & \textbf{37.94} & \textbf{36.24} & \textbf{36.04} & \textbf{34.24} \\ 
                    \midrule
\multirow{2}{*}{11} & AT-BSL    & 67.89          & 39.78          & 35.21          & 33.15          & 33.20          & 31.57          \\
                    & AT-BSL-RA & \textbf{68.46} & \textbf{42.10} & \textbf{37.63} & \textbf{34.58} & \textbf{34.26} & \textbf{33.12} \\ \midrule
\multirow{2}{*}{13} & AT-BSL    & \textbf{69.07} & 39.82          & 35.19          & 33.12          & 32.91          & 31.44          \\
                    & AT-BSL-RA & 68.90          & \textbf{42.18} & \textbf{37.89} & \textbf{34.93} & \textbf{34.58} & \textbf{33.35} \\
                    \bottomrule[1.0pt]
\end{tabular}}
\end{center}
\vspace{-0.15 in}
\end{table}

\subsection{Different Weight Decay}

\begin{figure}[t]
    \centering
    \begin{tabular}{cc}
        \includegraphics[width=0.45\linewidth]{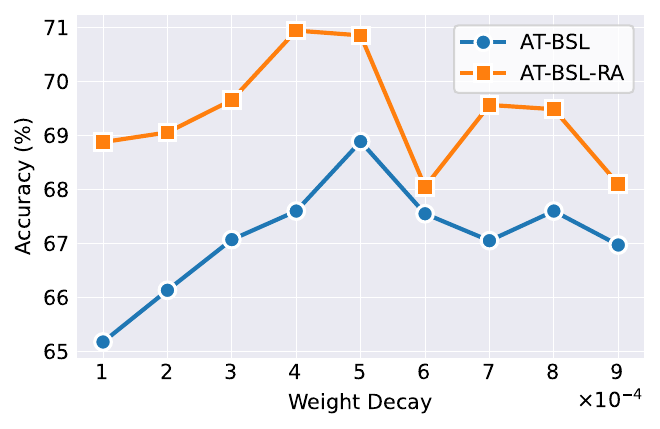} &
        \includegraphics[width=0.45\linewidth]{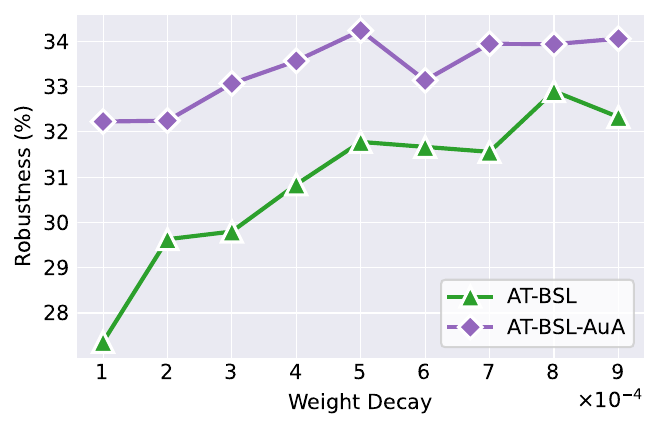} \\
        (a)& (b)
    \end{tabular}
    \vspace{-0.1 in}
    \caption{The clean accuracy and robustness under AA for various algorithms with different weight decay using ResNet-18 on CIFAR-10-LT at the best checkpoint.}
    \label{fig_different_wd}
    \vspace{-0.2 in}
\end{figure}

During our replication of the experiments of REAT~\cite{li2023adversarial}, we observe a discrepancy in the weight decay parameters used: REAT employed a weight decay of $5\times10^{-4}$, contrasting with $2\times10^{-4}$ used by RoBal~\cite{wu2021adversarial}. This leads us to conduct experiments using varying values of weight decay. The results, depicted in Fig. \ref{fig_different_wd}, indicate that a weight decay of $5\times10^{-4}$ offers a significant improvement over $2\times10^{-4}$ in terms of both accuracy and robustness. However, further increasing the weight decay beyond $5\times10^{-4}$ results in a noticeable decline in accuracy. Therefore, we employ a weight decay of $5\times10^{-4}$ in our experiments.

\subsection{Different Batch Sizes}

While replicating the experiments of REAT~\cite{li2023adversarial}, we note an inconsistency in the batch size settings: REAT utilized a batch size of 128, whereas RoBal utilized 64. To address this, we conduct experiments with different batch sizes, and the results are presented in Table \ref{tab_different_bs}. The findings indicate that the performance with batch sizes 64 and 128 are comparable, and both outperform larger batch sizes; however, 128 is more commonly used and helps speed up training. Consequently, we employ a batch size of 128 in our experiments.

\begin{table}[t]
\begin{center}
\caption{The robustness for various algorithms with different batch sizes using ResNet-18 on CIFAR-10-LT at the best checkpoint. Better results are \textbf{bolded}. BS: Batch Size.}
\label{tab_different_bs}
\vspace{-0.1 in}
\setlength{\tabcolsep}{0.7mm}{\begin{tabular}{cccccccc}
\toprule
BS                   & Method    & Clean          & FGSM           & PGD            & CW             & LSA            & AA             \\ \midrule
\multirow{2}{*}{64}  & AT-BSL    & \textbf{67.82} & 41.42          & 36.57          & 34.41          & 34.22          & 32.67          \\
                     & AT-BSL-RA & 66.70          & \textbf{41.85} & \textbf{37.87} & \textbf{35.34} & \textbf{34.83} & \textbf{33.78} \\ \midrule
\multirow{2}{*}{128} & AT-BSL    & 68.89          & 40.08          & 35.27          & 33.47          & 33.46          & 31.78          \\
                     & AT-BSL-RA & \textbf{70.86} & \textbf{43.06} & \textbf{37.94} & \textbf{36.24} & \textbf{36.04} & \textbf{34.24} \\ \midrule
\multirow{2}{*}{256} & AT-BSL    & 66.72          & 37.66          & 33.08          & 31.55          & 31.42          & 29.98          \\
                     & AT-BSL-RA & \textbf{67.93} & \textbf{41.05} & \textbf{36.60} & \textbf{33.78} & \textbf{33.43} & \textbf{31.97} \\ \midrule
\multirow{2}{*}{512} & AT-BSL    & 60.01          & 35.45          & 32.27          & 29.44          & 29.02          & 28.15          \\
                     & AT-BSL-RA & \textbf{63.25} & \textbf{37.66} & \textbf{34.38} & \textbf{31.14} & \textbf{30.59} & \textbf{29.53} \\ \bottomrule[1.0pt]
\end{tabular}}
\end{center}
\vspace{-0.1 in}
\end{table}

\subsection{Different Training Epochs}
As indicated in Table \ref{tab_different_epochs}, without data augmentation, the results between 80 and 100 training epochs show little difference. However, with data augmentation, we observe that a higher number of training epochs leads to increased robustness. This improvement is likely attributable to the augmented diversity of examples, necessitating a more extended learning period for the model.

\begin{table}[t]
\begin{center}
\caption{The robustness for various algorithms with different training epochs using ResNet-18 on CIFAR-10-LT at the best checkpoint. Better results are \textbf{bolded}.}
\label{tab_different_epochs}
\vspace{-0.1 in}
\setlength{\tabcolsep}{0.8mm}{\begin{tabular}{ccccccc}
\toprule[1.0pt]
Method       & Clean          & FGSM           & PGD            & CW             & LSA            & AA             \\ \midrule
AT-BSL-80    & 66.68          & \textbf{40.18} & \textbf{36.11} & \textbf{33.87} & \textbf{33.64} & \textbf{31.95} \\
AT-BSL       & \textbf{68.89} & 40.08          & 35.27          & 33.47          & 33.46          & 31.78          \\ \midrule
AT-BSL-RA-80 & 69.39          & 41.93          & 37.20          & 34.82          & 34.36          & 32.92          \\
AT-BSL-RA    & \textbf{70.86} & \textbf{43.06} & \textbf{37.94} & \textbf{36.24} & \textbf{36.04} & \textbf{34.24} \\ \bottomrule[1.0pt]
\end{tabular}}
\end{center}
\vspace{-0.2 in}
\end{table}

\subsection{Hyperparameter Tuning of RoBal}
Through hyperparameter tuning similar to those done for AT-BSL using ResNet-18 on CIFAR-10-LT, we find that RoBal achieve the best results with PGD-10, weight decay of $2\times10^{-4}$, batch size of 64, epochs of 60, and $\tau_b=1.5$. The robustness under AA reach 31.61\%, which is close to the performance of AT-BSL.

\subsection{Retraining RoBal and REAT}

Compared to RoBal~\cite{wu2021adversarial}, our primary experiments employ different experimental settings, including previously discussed variables like PGD steps, weight decay, batch size, and training epochs. To facilitate a fairer comparison, we adapt these settings in our main experiments: changing PGD-5 to PGD-10, weight decay from $2 \times 10^{-4}$ to $5 \times 10^{-4}$, batch size from 64 to 128, and increasing training epochs from 80 to 100, and then we retrain RoBal under these settings, referred to as RoBal (retraining). Compared with REAT~\cite{li2023adversarial}, the only discrepancy is in the training epochs. Therefore, we adjusted REAT's training epochs to 100 and conduct a retraining called REAT (retraining). The results are presented in Table \ref{tab_retraining_robal_and_reat}. The retrained RoBal is observed to achieve improved robustness, albeit at a slight cost to accuracy. Conversely, the retrained REAT displays even lower robustness than its initial version. Through this comparison, we note that the robustness achieved by the retrained RoBal and REAT is similar to that of the vanilla AT-BSL.

\begin{table}[t]
\begin{center}
\caption{The robustness for various algorithms using ResNet-18 on CIFAR-10-LT at the best checkpoint. The best results are \textbf{bolded}.}
\label{tab_retraining_robal_and_reat}
\vspace{-0.1 in}
\setlength{\tabcolsep}{0.5mm}{\begin{tabular}{ccccccc}
\toprule[1.0pt]
Method               & Clean          & FGSM           & PGD            & CW             & LSA            & AA             \\ \midrule
RoBal                & 70.34          & 40.50          & 35.93          & 31.05          & 31.10          & 29.54          \\
RoBal   (retraining) & 67.46          & 41.61          & \textbf{38.04} & 32.75          & 33.08          & 31.26          \\
REAT                 & 67.38          & 40.13          & 35.83          & 33.88          & 33.66          & 32.20          \\
REAT   (retraining)  & 67.38          & 39.51          & 35.15          & 33.53          & 33.31          & 31.77          \\
AT-BSL               & 68.89          & 40.08          & 35.27          & 33.47          & 33.46          & 31.78          \\
AT-BSL-RA            & \textbf{70.86} & \textbf{43.06} & 37.94          & \textbf{36.24} & \textbf{36.04} & \textbf{34.24} \\ \bottomrule[1.0pt]
\end{tabular}}
\end{center}
\vspace{-0.1 in}
\end{table}

\subsection{Other Data Augmentations}

\noindent \textbf{Data Augmentations Designed for Long-Tailed Recognition.}
CUDA~\cite{ahn2023cuda}\footnote{\href{https://github.com/sumyeongahn/cuda_ltr}{https://github.com/sumyeongahn/cuda\_ltr}} initially explored the relationship between the degree of augmentation and class performance, discovering that an appropriate level of augmentation needs to be allocated for each class to mitigate class imbalance issues. Inspired by this finding, ~\cite{ahn2023cuda} introduces a simple yet efficient novel curriculum to identify the appropriate data augmentation strength for each class, called CUDA: CUrriculum of Data Augmentation for long-tailed recognition. To assess CUDA's performance in adversarial training under long-tailed distributions, we augment AT-BSL with CUDA, referred to as AT-BSL-CUDA, and compared it with the vanilla AT-BSL, as shown in the Table \ref{tab_other_aug}. The results suggest that CUDA's performance in adversarial training under long-tailed distributions appears less effective than RA.

\noindent \textbf{Data Augmentations Designed for Adversarial Training.} DAJAT~\cite{dajat}\footnote{\href{https://github.com/val-iisc/dajat}{https://github.com/val-iisc/dajat}} proposes a data augmentation technique designed explicitly for adversarial training. ~\cite{dajat} initially conceptualizes data augmentation as a domain generalization problem during the training process. Subsequently, they introduce Diverse Augmentation-based Joint Adversarial Training (DAJAT), effectively integrating data augmentation into adversarial training. Since DAJAT's experiments are based on TRADES~\cite{zhang2019theoretically}, it cannot be directly applied to augment AT-BSL. We conduct comparative analyses between vanilla TRADES and DAJAT. The comparison in Table \ref{tab_other_aug} reveals that DAJAT still contributes to improved robustness in long-tailed adversarial training, showing comparable effectiveness to TRADES-RA.

IDBH~\cite{idbh}\footnote{\href{https://github.com/treelli/da-alone-improves-at}{https://github.com/treelli/da-alone-improves-at}} is another data augmentation technique that is specifically formulated for adversarial training. ~\cite{idbh} discovers that the diversity and hardness of data augmentation play a crucial role in combating adversarial overfitting. Overall, diversity enhances both accuracy and robustness, while hardness can improve robustness to a certain extent, but at the expense of accuracy and beyond a certain threshold, it diminishes both. ~\cite{idbh} introduces a novel cropping transformation method called Cropshift to mitigate robust overfitting. Building on Cropshift, ~\cite{idbh} proposes a new augmentation scheme called Improved Diversity and Balanced Hardness (IDBH). We utilize IDBH to augment AT-BSL, referred to as AT-BSL-IDBH. Upon comparison in Table \ref{tab_other_aug}, it is found that IDBH's effectiveness is less pronounced than RA on long-tailed datasets.

\begin{table}[t]
\begin{center}
\caption{The robustness for various algorithms with different data augmentations using ResNet-18 on CIFAR-10-LT at the best checkpoint. The best results are \textbf{bolded}.}
\label{tab_other_aug}
\vspace{-0.1 in}
\setlength{\tabcolsep}{0.8mm}{\begin{tabular}{ccccccc}
\toprule[1.0pt]
Method        & Clean          & FGSM           & PGD            & CW             & LSA            & AA             \\ \midrule
TRADES        & 43.61          & 29.18          & \textbf{27.81} & \textbf{26.73} & \textbf{26.58} & \textbf{26.41}          \\
DAJAT         & 42.04          & \textbf{29.34} & 27.70          & 26.47          & 26.36          & 26.27          \\
TRADES-RA & \textbf{44.45} & 29.18          & 27.61          & 26.51          & 26.47          & 26.27          \\ \midrule
AT-BSL        & 68.89          & 40.08          & 35.27          & 33.47          & 33.46          & 31.78          \\
AT-BSL-CUDA   & 68.05          & 40.06          & 36.48          & 33.07          & 32.75          & 31.49          \\
AT-BSL-IDBH   & 70.80          & 39.54          & 33.30          & 32.56          & 33.01          & 31.24          \\
AT-BSL-RA & \textbf{70.86} & \textbf{43.06} & \textbf{37.94} & \textbf{36.24} & \textbf{36.04} & \textbf{34.24} \\ \bottomrule[1.0pt]
\end{tabular}}
\end{center}
\vspace{-0.15 in}
\end{table}

\subsection{Using Data Generated by Diffusion Models}

To investigate the potential of leveraging data generated by diffusion models to improve the robustness of AT-BSL, we train a diffusion model, DDPM++, for CIFAR-10-LT, selecting the version with the best Fréchet Inception Distance (FID) of 6.92 after 18 sampling steps following ~\cite{edm, better}. For the generation of 1 million data points, we produce 100,000 images per class, culminating in a total of 1 million images. Following ~\cite{better}, we set the proportion of unsupervised data to 0.7 and train a ResNet-18 using AT-BSL, which we refer to as AT-BSL-DM. The results presented in Table \ref{tab_edm} clearly demonstrate the significant improvement in robustness afforded by incorporating data generated by diffusion models.

\begin{table}[t]
\begin{center}
\caption{The robustness for various algorithms with different training epochs using ResNet-18 on CIFAR-10-LT at the best checkpoint. Better results are \textbf{bolded}.}
\label{tab_edm}
\vspace{-0.1 in}
\setlength{\tabcolsep}{0.8mm}{\begin{tabular}{ccccccc}
\toprule[1.0pt]
Method    & Clean          & FGSM           & PGD            & CW             & LSA            & AA             \\ \midrule
AT-BSL    & 68.89          & 40.08          & 35.27          & 33.47          & 33.46          & 31.78          \\
AT-BSL-RA & 70.86          & 43.06          & 37.94          & 36.24          & 36.04          & 34.24          \\
AT-BSL-DM & \textbf{72.61} & \textbf{47.09} & \textbf{42.01} & \textbf{41.56} & \textbf{41.89} & \textbf{39.48} \\ \bottomrule[1.0pt]
\end{tabular}}
\end{center}
\vspace{-0.1 in}
\end{table}

\subsection{Different Adversarial Training Methods}

To further validate the hypothesis that data augmentation alone improves robustness under long-tailed distributions, we conduct experiments across various adversarial training methods, employing AuA or RA. As evidenced in Table \ref{tab_different_at}, with few exceptions, data augmentation is beneficial for robustness. This effect is particularly common on CIFAR-100-LT, likely due to the reduced number of training examples per class in this dataset, leading to a more substantial reliance on data augmentation techniques.

\begin{table*}[t]
\begin{center}
\caption{The robustness for various algorithms with/without data augmentations at the best checkpoint. On the combination of ResNet-18 and CIFAR-10-LT, RA is employed, whereas, for other model and dataset combinations, AuA is utilized. Better results are \textbf{bolded}.}
\label{tab_different_at}
\vspace{-0.1 in}
\setlength{\tabcolsep}{1.9mm}{\begin{tabular}{ccccccccccccc}
\toprule[1.0pt]
\multirow{3}{*}{Method} & \multicolumn{6}{c}{CIFAR-10-LT}                                                                     & \multicolumn{6}{c}{CIFAR-100-LT}                                                                    \\ \cmidrule(l){2-7} \cmidrule(l){8-13} 
                        & \multicolumn{3}{c}{ResNet-18}                     & \multicolumn{3}{c}{WideResNet-34-10}             & \multicolumn{3}{c}{ResNet-18}                     & \multicolumn{3}{c}{WideResNet-34-10}             \\ \cmidrule(l){2-4} \cmidrule(l){5-7} \cmidrule(l){8-10} \cmidrule(l){11-13} 
                        & Clean          & PGD            & AA             & Clean          & PGD            & AA             & Clean          & PGD            & AA             & Clean          & PGD            & AA             \\ \midrule
AT~\cite{madry2018towards}                      & \textbf{49.35} & 27.30          & 25.76          & 59.21          & 27.88          & 27.07          & 41.20          & 14.59          & 13.62          & 45.18          & 16.36          & 15.60          \\
AT-RA/AuA               & 44.31          & \textbf{27.81} & \textbf{25.90} & \textbf{62.98} & \textbf{33.40} & \textbf{31.64} & \textbf{45.17} & \textbf{19.78} & \textbf{17.22} & \textbf{50.00} & \textbf{21.87} & \textbf{19.44} \\ \midrule
TRADES~\cite{zhang2019theoretically}                  & 43.61          & \textbf{27.81} & \textbf{26.41} & 51.28          & 28.70          & 27.72          & 38.12          & 17.89          & 15.59          & 41.71          & 19.85          & 17.91          \\
TRADES-RA/AuA           & \textbf{44.45} & 27.61          & 26.27          & \textbf{55.89} & \textbf{31.53} & \textbf{29.77} & \textbf{42.14} & \textbf{19.69} & \textbf{16.12} & \textbf{46.23} & \textbf{22.78} & \textbf{19.52} \\ \midrule
MART~\cite{wang2019improving}                    & \textbf{48.61} & \textbf{30.29} & \textbf{27.73} & \textbf{49.13} & \textbf{32.32} & \textbf{29.60} & \textbf{38.46} & 21.36          & 17.51          & 41.32          & 23.27          & 19.98          \\
MART-RA/AuA             & 43.76          & 29.86          & 26.77          & 48.07          & 31.93          & 28.31          & 38.01          & \textbf{22.64} & \textbf{18.68} & \textbf{43.43} & \textbf{25.41} & \textbf{21.26} \\ \midrule
AWP~\cite{wu2020adversarial}                     & \textbf{49.29} & \textbf{31.20} & \textbf{29.53} & \textbf{50.91} & \textbf{31.85} & \textbf{30.06} & \textbf{41.53} & 21.79          & 18.61          & \textbf{45.66} & 23.88          & 20.56          \\
AWP-RA/AuA              & 45.28          & 30.56          & 28.73          & 44.06          & 29.91          & 27.81          & 41.07          & \textbf{23.02} & \textbf{19.37} & 45.27          & \textbf{25.76} & \textbf{21.60} \\ \midrule
GAIRAT~\cite{zhang2020geometry}                  & \textbf{50.83} & \textbf{27.46} & \textbf{20.41} & 59.89          & 30.40          & 25.38          & 38.99          & 18.05          & \textbf{15.61} & 36.41          & 17.31          & 14.77          \\
GAIRAT-RA/AuA           & 43.56          & 27.34          & 17.82          & \textbf{66.43} & \textbf{37.96} & \textbf{25.53} & \textbf{41.94} & \textbf{19.18} & 14.82          & \textbf{49.75} & \textbf{22.19} & \textbf{18.24} \\ \midrule
LAS-AT~\cite{jia2022adversarial}                  & \textbf{52.81} & 30.32          & 28.53          & 57.52          & 29.86          & 28.84          & 44.33          & 19.59          & 16.15          & 45.86          & 20.02          & 17.35          \\
LAS-AT-RA/AuA           & 51.20          & \textbf{31.20} & \textbf{29.18} & \textbf{59.14} & \textbf{34.51} & \textbf{32.54} & \textbf{45.18} & \textbf{22.78} & \textbf{18.61} & \textbf{49.73} & \textbf{24.09} & \textbf{20.79} \\ \midrule
RoBal~\cite{wu2021adversarial}                   & \textbf{70.34} & 35.93          & 29.54          & \textbf{72.82} & 36.42          & 30.49          & 45.93          & 17.40          & 16.42          & 50.08          & 18.84          & 17.90          \\
RoBal-RA/AuA            & 68.66          & \textbf{37.50} & \textbf{30.06} & 72.57          & \textbf{40.54} & \textbf{31.87} & \textbf{47.75} & \textbf{19.93} & \textbf{18.04} & \textbf{54.12} & \textbf{21.41} & \textbf{19.66} \\ \midrule
REAT~\cite{li2023adversarial}                    & \textbf{67.38} & 35.83          & \textbf{32.20} & \textbf{73.16} & 35.94          & 33.20          & 46.28          & 18.85          & 16.54          & 50.29          & 20.82          & 18.65          \\
REAT-RA/AuA             & 66.64          & \textbf{36.97} & 31.84          & 72.05          & \textbf{40.05} & \textbf{35.74} & \textbf{47.65} & \textbf{22.86} & \textbf{18.48} & \textbf{50.10} & \textbf{25.07} & \textbf{20.81} \\ \midrule
AT-BSL                  & 68.89          & 35.27          & 31.78          & 73.19          & 35.60          & 32.80          & 45.59          & 18.05          & 15.97          & 50.04          & 19.66          & 18.04          \\
AT-BSL-RA/AuA           & \textbf{70.86} & \textbf{37.94} & \textbf{34.24} & \textbf{75.17} & \textbf{40.84} & \textbf{37.15} & \textbf{48.39} & \textbf{22.96} & \textbf{18.90} & \textbf{53.08} & \textbf{25.40} & \textbf{21.43} \\ \bottomrule[1.0pt]
\end{tabular}}
\end{center}
\end{table*}

\subsection{Standard Deviation}
We repeat AT-BSL and AT-BSL-RA five times using ResNet-18 on CIFAR-10-LT. Their mean and standard deviation of robustness under AA are $31.65 \pm 0.45$ and $34.12\pm0.51$, respectively. The relatively small variance indicates the stability of our training process.

\subsection{Computational Cost Comparison}
In this section, we compare the computational costs of AT-BSL and AT-BSL-RA/AuA regarding average training time per epoch and GPU memory usage. The detailed results are summarized in Table \ref{tab_cost}. The comparison indicates that the introduction of data augmentation incurs a negligible increase in time cost without imposing additional memory overhead.

\begin{table*}[t]
\begin{center}
\caption{The time and memory for various algorithms. On the combination of ResNet-18 and CIFAR-10-LT, RA is employed, whereas, for other model and dataset combinations, AuA is utilized. All experiments are run on NVIDIA RTX 3090.}
\label{tab_cost}
\begin{tabular}{cccccc}
\toprule[1.0pt]
\multirow{2}{*}{Dataset}      & \multirow{2}{*}{Method} & \multicolumn{2}{c}{ResNet-18}             & \multicolumn{2}{c}{WideResNet-34-10}      \\ \cmidrule(l){3-6} 
                              &                         & Time (s)                  & Memory (MiB) & Time (s)                   & Memory (MiB) \\ \midrule
\multirow{2}{*}{CIFAR-10-LT}  & AT-BSL                  & 22.37 & 1345         & 200.70                     & 4293         \\
                              & AT-BSL-RA/AuA           &22.43 & 1345         &201.42 & 4293         \\ \midrule
\multirow{2}{*}{CIFAR-100-LT} & AT-BSL                  & 30.94                     & 1347         & 277.82                     & 4293         \\
                              & AT-BSL-RA/AuA           & 31.25                     & 1347         & 279.66                     & 4293         \\ \bottomrule[1.0pt]
\end{tabular}
\end{center}
\end{table*}

\section{Comparison with Concurrent Works}

Concurrently and independently from our work, REAT~\cite{li2023adversarial} has also explored adversarial training under long-tailed distributions. ~\cite{li2023adversarial} identifies that compared to conventional adversarial training on balanced datasets, this process tends to produce imbalanced adversarial examples and feature embedding spaces, resulting in reduced robustness on tail data. To address this issue, ~\cite{li2023adversarial} introduces a novel adversarial training framework: Re-balancing Adversarial Training (REAT). This framework comprises two key components: (1) a new training strategy inspired by the concept of effective numbers, guiding the model to generate more balanced and informative adversarial examples, and (2) a meticulously designed penalty function aimed at enforcing a satisfactory feature space. Notably, the experimental settings utilized in our paper are fundamentally consistent with those employed in REAT. Moreover, as shown in Tables \ref{tab_res_cifar10}, \ref{tab_wrn_cifar10}, \ref{tab_res_cifar100}, and \ref{tab_wrn_cifar100}, the robustness achieved by our implemented vanilla AT-BSL is comparable to that of REAT.

\end{document}